\documentclass{article}

\usepackage{microtype}
\usepackage{graphicx}
\usepackage{subcaption}
\usepackage{booktabs} % for professional tables

\usepackage{hyperref}

\usepackage[preprint]{icml2026}

\usepackage{amsmath}
\usepackage{amssymb}
\usepackage{mathtools}
\usepackage{amsthm}

\usepackage{algorithm}
\usepackage{algorithmic}

\usepackage{multirow}
\usepackage{makecell}
\usepackage[skins]{tcolorbox}
\usepackage{tikz}

\usepackage[table]{xcolor}
\usepackage{longtable}

\usepackage{array}           % 增强表格功能
\usepackage{tabularx}        % 自动调整列宽
\usepackage{colortbl}        % 表格单元格着色
\usepackage{pifont}          % 特殊符号 (✓ ✗)
\newcommand{\cmark}{\ding{51}}   % ✓
\newcommand{\xmark}{\ding{55}}

\usepackage[capitalize,noabbrev]{cleveref}

\theoremstyle{plain}

\theoremstyle{definition}

\theoremstyle{remark}

\usepackage[textsize=tiny]{todonotes}

\icmltitlerunning{}
\makeatletter
\def\headrulewidth{0pt}
\makeatother
\begin{document}

\twocolumn[
  \icmltitle{HuggingR$^{4}$: A Progressive Reasoning Framework for Discovering \\ Optimal Model Companions}



  \icmlsetsymbol{equal}{*}

  \begin{icmlauthorlist}
    \icmlauthor{Shaoyin Ma}{equal,yyy}
    \icmlauthor{Chenggong Hu}{equal,yyy}
    \icmlauthor{Huiqiong Wang}{yyy}
    \icmlauthor{Li Sun}{yyy}
    \icmlauthor{Mingli Song}{yyy}
    \icmlauthor{Jie Song}{yyy}
  \end{icmlauthorlist}

  \icmlaffiliation{yyy}{Zhejiang University, Hangzhou, China}

  \icmlcorrespondingauthor{Jie Song}{sjie@zju.edu.cn}


  \vskip 0.3in
]



\printAffiliationsAndNotice{\icmlEqualContribution}

\begin{abstract}
  Building effective LLM agents increasingly requires selecting appropriate AI models as tools from large open repositories (\textit{e.g.}, HuggingFace with \textgreater 2M models) based on natural language requests. Unlike invoking a fixed set of API tools, repository-scale model selection must handle massive, evolving candidates with incomplete metadata. Existing approaches incorporate full model descriptions into prompts, resulting in prompt bloat, excessive token costs, and limited scalability. To address these issues, we propose \textbf{HuggingR$^{4}$}, the first framework to recast model selection as an iterative reasoning process rather than one-shot retrieval. By synergistically integrating \textbf{R}easoning, \textbf{R}etrieval, \textbf{R}efinement, and \textbf{R}eflection, HuggingR$^4$ progressively decomposes user intent, retrieves candidates through multi-round deliberation, refines selections via fine-grained analysis, and validates results through reflection. To facilitate rigorous evaluation, we introduce a large-scale benchmark comprising 14,399 diverse user requests across 37 task categories. Experiments demonstrate that HuggingR$^4$ achieves $92.03\%$ workability and $82.46\%$ reasonability—outperforming current state-of-the-art baselines by $26.51\%$ and $33.25\%$, respectively, while reducing token consumption by $6.9 \times$.
\end{abstract}

\section{Introduction}

The Model Context Protocol (MCP) has catalyzed the evolution of large language models by making the integration of diverse external interfaces a key paradigm in AI agent development, with remarkable results across various fields \cite{zhang2025multi, contreras2025multiclass, kadiyala2024implementation}. These external interfaces can fall into two categories: functional APIs \cite{li2023api, mastouri2025making, liu2025rescueadi}, and system-level interfaces \cite{ehtesham2025survey, ahmadi2025mcp}. By leveraging these interfaces, LLMs extend beyond pure text-based reasoning, enabling them to process visual inputs, access real-time data, and perform complex tasks. As the interface ecosystem rapidly expands across vision and cross-modal domains, selecting the most appropriate interfaces from vast repositories based on user queries has become a critical challenge in building effective AI agents.

\begin{figure}[t]
\centering
\vspace{5pt}
\includegraphics[width=1.0\columnwidth]{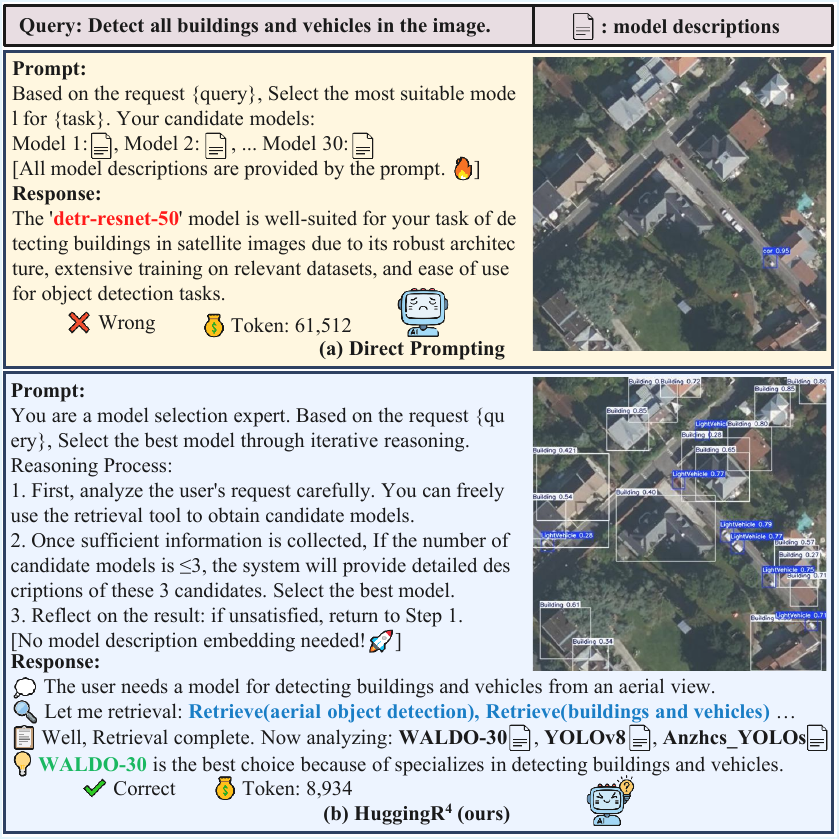}
\caption{Comparison of baseline and HuggingR$^{4}$ prompts and responses. For the same query, HuggingR$^{4}$ chooses the correct model with 6.9$\times$ fewer tokens, whereas the baseline chooses an incorrect model. Right: execution results.}
\label{fig:1}
\vspace{-20pt}
\end{figure}

Among these interfaces, community-driven AI models represent a rich yet underexplored resource. While existing work has primarily focused on invoking a fixed set of curated APIs \cite{gan2025rag, lumer2025scalemcp}, the selection of appropriate models from large, evolving community repositories remains largely unexplored. To date, the HuggingFace community has amassed over 2.5 million models, 776k datasets, and supports 4,700 languages, expanding the available tool set for AI agents. We believe that each model excels in its specialized domain, and models fine-tuned on domain-specific data are more adept at meeting users' particular needs. For instance, BioBERT \cite{lee2020biobert} is better suited for medical text analysis than BERT \cite{koroteev2021bert}. However, due to the rapid expansion of the community model ecosystem and metadata gaps, retrieval system performance is impacted. Therefore, developing model selection algorithms that balance both efficiency and accuracy has become a critical bottleneck in unlocking the full potential of community models.

% Initial community model selection approaches, such as HuggingGPT \cite{shen2023hugginggpt}, achieved this by directly embedding model descriptions into prompts, leading to significant token consumption. Recent research has shifted towards a paradigm driven by meta-learners, in which models are retrieved from multiple dimensions based on user input \cite{li2023automrm, zhou2024you, lu2023content}. Furthermore, Modelgalaxy \cite{zhang2024modelgalaxy} introduced retrieval-augmented generation (RAG) \cite{lewis2020retrieval}, where an LLM-driven query rewriting mechanism optimizes retrieval queries to fetch candidate models from a vector database. However, these approaches fail to perform fine-grained analysis of model descriptions. This makes it difficult to achieve an optimal match between the heterogeneous nature of community models and the diverse needs of users.

Current mainstream approaches for community model selection, 
such as HuggingGPT~\cite{shen2023hugginggpt}, achieve model 
selection by directly embedding all candidate model descriptions 
into prompts, leading to significant token consumption 
(Figure~\ref{fig:1}). Recent research has shifted towards a paradigm driven by meta-learners that select models based on dataset similarity \cite{li2023automrm, zhou2024you, lu2023content, zhang2024modelgalaxy}. However, these methods are limited to fixed, small-scale, task-specific repositories and fail to generalize to other tasks or adapt to dynamically evolving large-scale model communities. Moreover, existing approaches fail to perform fine-grained analysis of model descriptions. This makes it difficult to achieve an optimal match between the heterogeneous nature of community models and the diverse needs of users.

To address these challenges, we introduce HuggingR$^4$, a novel reasoning-driven framework for scalable model selection in dynamic community repositories. It employs a coarse-to-fine strategy that combines efficient broad-scale retrieval with precise fine-grained refinement, achieving state-of-the-art performance while reducing token consumption by 85.6\% (Figure~\ref{fig:1}). HuggingR$^4$ construction process comprises three stages: \textbf{(1) Iterative Reasoning and Dual-Stream Retrieval.} The LLM performs step-by-step reasoning to decompose user queries and iteratively retrieves a compact set of candidates. \textbf{(2) Fine-Grained Semantic Refinement.} The LLM conducts fine-grained analysis of candidates' full descriptions to select the most suitable model based on task-specific requirements. \textbf{(3) Meta-Cognitive Reflection.} The LLM validates selections through self-reflection and dynamically adjusts retrieval parameters if necessary. To handle pervasive metadata gaps, we integrate a failure traceback mechanism after each retrieval operation. Additionally, we introduce a novel sliding window strategy that intelligently manages context window usage, optimizing both information coverage and token efficiency.

To evaluate our approach, we introduce \textbf{ModelSelect-Bench}, the first large-scale benchmark for repository-scale model selection. It comprises 1,016 single-task and 13,383 multi-task requests spanning vision, language, audio, and multimodal scenarios, each rigorously human-annotated with optimal model selections. We employ two complementary metrics: \textit{workability} (whether the selected model executes successfully), and \textit{reasonability} (whether outputs meet user requirements). In summary, our contributions are:

% \vspace{-10pt}
\setlength{\leftmargini}{10pt}
\begin{itemize}
\setlength{\itemsep}{-0.3em}
    \item We propose \textbf{HuggingR$^4$}, the first scalable framework for model selection in community repositories, combining iterative LLM reasoning with advanced retrieval techniques to scale seamlessly to large, evolving repositories.
    
    \item We construct \textbf{ModelSelect-Bench}, the first comprehensive global benchmark comprising 14K+ requests across vision, language, audio, and multimodal domains.
    
    \item Extensive experiments demonstrate that HuggingR$^4$ achieves SOTA performance across multiple LLM architectures while reducing token consumption by $6.9 \times$.
\end{itemize}
\vspace{-10pt}
\section{Related Work}
\label{sec:Related_Work}

% Model selection has become a critical step in building LLM agents, aiming to identify the most suitable model from a pool of candidates for a given task. While not a new research problem, early work primarily relied on rule-based heuristics and classical machine learning methods \cite{taylor2018adaptive,marco2020optimizing} to select neural models for specific. Then, the advent of deep learning led to the development of lightweight selector models \cite{wang2017automated,meng2021measuring,hu2023laf} and meta-learning techniques \cite{li2021meta,luo2020metaselector,wei2025efficient} for adaptive model selection. However, these approaches were typically designed for small, static model sets, limiting their scalability to large and dynamic repositories such as HuggingFace. As the number of available models continues to grow, more recent work has shifted towards using user input as the basis for model selection \cite{bolya2021scalable,lu2023content,li2023automrm,zhou2024you,wanaskar2025multimodal}. HuggingGPT \cite{shen2023hugginggpt} demonstrated the feasibility of using LLMs to invoke HuggingFace models by embedding model descriptions directly into prompts. ModelGalaxy \cite{zhang2024modelgalaxy} further integrated prior efforts into a unified model selection platform. While these approaches support basic retrieval and invocation, they often fail to account for user preferences or fine-grained task requirements, thus limiting their effectiveness in identifying optimal models.

\begin{figure*}[htb]
\centering
\includegraphics[width=1.0\textwidth]{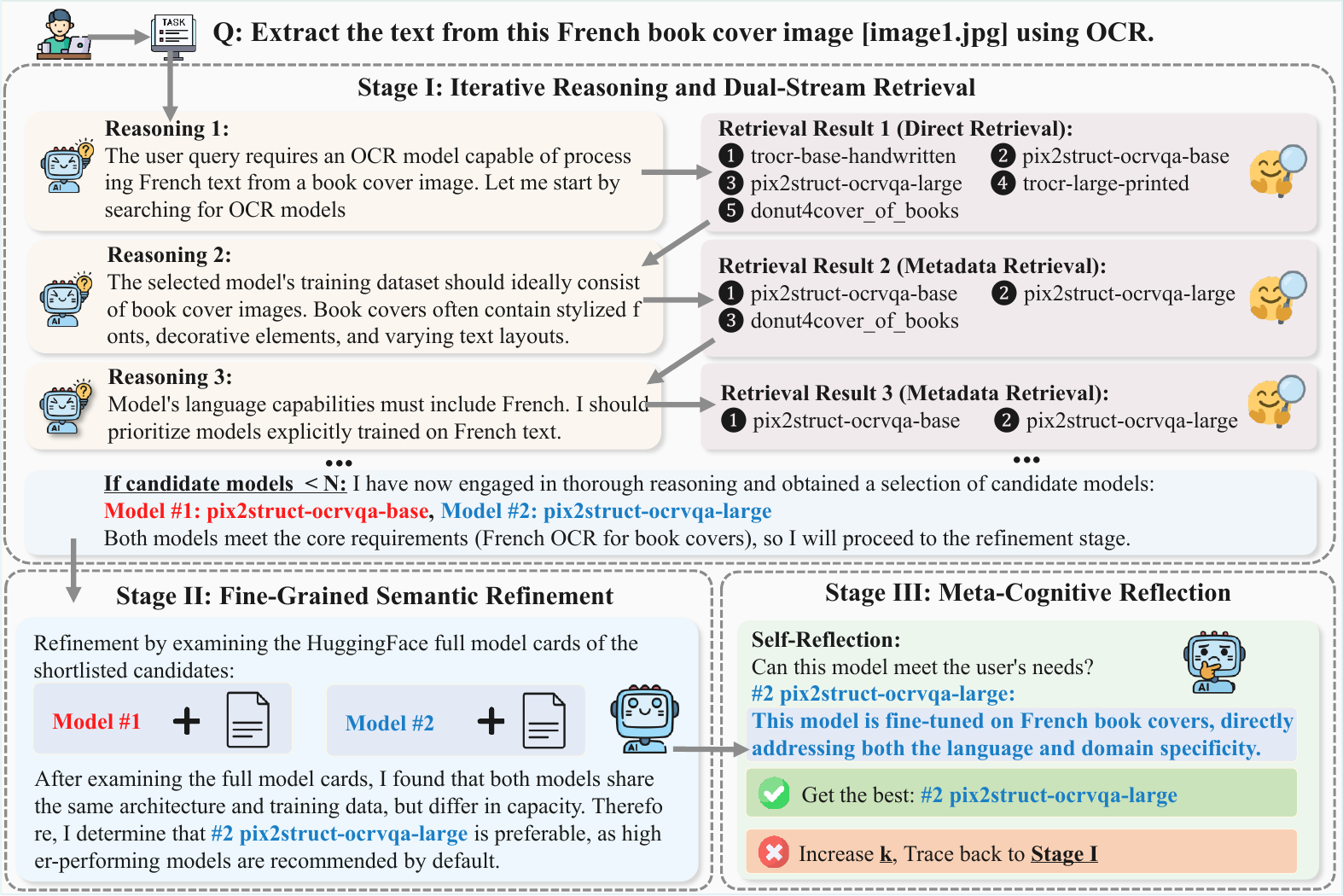}
% \vspace{-15pt}
\caption{Workflow of HuggingR$^4$ with an example. The process includes: 1) Reasoning and Retrieval: Iterative querying and top-k candidate retrieval from the vector database. 2) Refinement: Fine-grained model selection via sliding window access to complete model cards. 3) Reflection: Self-evaluation to verify model suitability.}

\label{fig:2}
\vspace{-15pt}
\end{figure*}

\paragraph{Model Selection.} 
Model selection aims to identify the most suitable model from a candidate pool for specific tasks. Existing approaches fall into two paradigms: data-driven and query-driven methods. 

Data-driven methods require access to datasets and extract meta-features for model recommendation. Early work relied on rule-based heuristics and classical machine learning \cite{taylor2018adaptive, marco2020optimizing}, later evolving into lightweight selector models \cite{wang2017automated, meng2021measuring, hu2023laf}. Recent advances focus on metadata-based \cite{li2021meta, luo2020metaselector, lu2023content, li2023automrm, zhou2024you, wanaskar2025multimodal, zhang2024modelgalaxy} 
and graph-based \cite{li2024model, wang2023selecting} approaches. However, these methods face limitations in community model scenarios: \textit{(1)} most HuggingFace models lack comprehensive dataset information, \textit{(2)} real-time dataset acquisition is infeasible, and \textit{(3)} they cannot handle diverse user preferences beyond dataset statistics.

Query-driven methods accept natural language queries as input, eliminating dataset dependency. HuggingGPT \cite{shen2023hugginggpt} pioneered this direction by embedding model descriptions into LLM prompts, but suffers from limited scalability: embedding all descriptions becomes impractical as repositories grow to tens of thousands of models, and lacks 
mechanisms to capture fine-grained task requirements (e.g., language-specific needs) or user preferences. More broadly, existing methods fail to address the unique challenges of community model selection: incomplete metadata, 
rapidly evolving repositories, and diverse tasks across vision, language, audio, and multimodal domains.

\paragraph{Tool Selection vs. Model Selection.}
Unlike prior work on API tool selection, which operates on discrete function categories (weather vs. email) with structured schemas, community model selection faces \textit{continuous capability spaces} where thousands of models (e.g., 6,117 sentiment models) differ subtly in domain and architecture. This transforms the problem from category-based classification to fine-grained ranking under incomplete metadata and context-dependent performance. We provide a detailed comparison of these paradigm differences in Appendix~\ref{app:vs}.

% \subsection{LLMs for Reasoning}
% \noindent\textbf{LLMs for Reasoning.} LLMs have demonstrated strong reasoning abilities, supporting advanced capabilities such as decision-making, planning, and task decomposition \cite{yao2023react, feng2025efficient}. To further enhance this capacity, various techniques have been proposed, including Chain-of-Thought Prompting \cite{wei2022chain, kojima2022large, wang2023boosting, kang2025c3ot}, symbolic and logical reasoning integration \cite{pan2023logic}, and optimization-based training strategies \cite{zhang2024rest, yu2025self, tong2024optimizing, yuan2025incentivizing}. In parallel, some studies have explored combining retrieval with reasoning, enabling LLMs to dynamically acquire external information at each step of the reasoning process and thereby support more adaptive, feedback-driven thinking \cite{huang2025improve, li2025search, wu2025agentic, xia2025improving, shi2025search, gerych2024knows}. Notably, when selecting appropriate tools, humans often follow an iterative process of thinking, retrieving, and rethinking \cite{riesbeck2013inside, kurtz2007converging, adair2016incorporating, zhao2023automatic} to arrive at better decisions. Inspired by this, we argue that selecting external model interfaces for LLM agents should likewise be viewed as a step-by-step process of reasoning and retrieval, rather than a single-step task.
\vspace{-10pt}
\paragraph{LLMs for Reasoning.} 
LLMs demonstrate strong reasoning through techniques like Chain-of-Thought prompting \cite{wei2022chain, kojima2022large, wang2023boosting, kang2025c3ot}, symbolic 
reasoning \cite{pan2023logic}, and optimization-based training \cite{zhang2024rest, yu2025self, tong2024optimizing, yuan2025incentivizing}. Recent work combines retrieval with reasoning, enabling LLMs to dynamically acquire information during multi-step processes \cite{huang2025improve, li2025search, wu2025agentic, xia2025improving, shi2025search, gerych2024knows}. Notably, when selecting appropriate tools, humans often follow an iterative process of thinking, retrieving, and rethinking \cite{riesbeck2013inside, kurtz2007converging, zhao2023automatic} to arrive at better decisions. Inspired by this, we argue that selecting external model interfaces for LLM agents should likewise be viewed as a step-by-step process, rather than a single-step task.
\vspace{-5pt}
\section{Methodology}
% \subsection{Problem Formulation}
% In this section, we first introduce the model retriever HuggingRA as the core component of HuggingRAR. And then we describe the workflow of the HuggingRAR framework, including its sliding window strategy.

\subsection{Task Formulation and Notation}
We consider the task of model selection within a massive, evolving ecosystem  (\textit{e.g.}, Hugging Face). Let $\mathcal{D} = \{m_1, m_2, ..., m_n\}$ denote a repository containing  candidate models, where $n$ is the number of models. Each model $m_i$ is characterized by a high-dimensional \textit{model card} $C(m_i)$, which we define as a composite tuple:
\begin{equation}
C(m_i) = \langle \mathcal{A}_{\text{base}}, \mathcal{A}_{\text{struct}}, \mathcal{T}_{\text{desc}} \rangle,
\end{equation}
where $\mathcal{A}_{\text{base}}$ represents atomic metadata (\textit{e.g.}, unique identifier, usage statistics, and canonical task labels), $\mathcal{A}_{\text{struct}}$ the schematized attributes (\textit{e.g.}, supported languages, dataset lineage, and licensing constraints), and $\mathcal{T}_{\text{desc}}$ unstructured semantic content (\textit{e.g.}, technical specifications, architectural nuances, and qualitative performance metrics).

Given a user query $q$ that encodes latent functional requirements and domain constraints, the objective is to find the optimal model companions $\mathcal{M}^*$. We frame this as a maximization of an alignment function $\Phi$:
\begin{equation}
    \mathcal{M}^* = \arg\max_{\mathcal{M} \subseteq \mathcal{D}} \Phi\left(q, C(\mathcal{M})\right).
\end{equation}
In the context of repository-scale selection, $\Phi$ is non-trivial to compute due to three primary challenges. (1) \textit{Informational Asymmetry}: the mapping between $q$ and $C(\mathcal{M})$ is often obscured by incomplete or noisy metadata in $\mathcal{A}_{\text{struct}}$. (2) \textit{Computational Tractability}: the cardinality of $\mathcal{D}$ precludes the direct application of LLM-based reasoning across all $\mathcal{T}_{\text{desc}}$. (3) \textit{Semantic Drift}: natural language requests $q$ often require iterative decomposition to align with the technical granularity of $C(\mathcal{M})$.

% In the HuggingFace ecosystem, each model is documented via a model card that serves as a standardized profile. A model card contains three primary types of information: (1) basic metadata including identifier, downloads, likes, and task designation; (2) structured metadata encompassing languages, datasets, licensing, and labels; and (3) descriptive content providing specifications, usage instructions, and performance characteristics.

% Formally, we define the model selection as follows: Given a user query $q$ and a candidate model repository $\mathcal{D} = \{m_1, m_2, ..., m_n\}$ where each model $m_i$ is represented by its corresponding model card, the objective is to identify the optimal model $m^* \in \mathcal{D}$ that best satisfies the user's specified needs.

% HuggingR$^4$ is a progressive reasoning framework designed to efficiently discover optimal model companions from large-scale model repositories. As illustrated in Figure~\ref{fig:2}, our approach consists of three stages: Reasoning and Retrieval, Refinement, Reflection. 
% Additionally, to efficiently manage computational resources and token consumption during this process, we introduce a novel sliding window strategy for model card access, as depicted in Figure 3. This strategy dynamically allocates different access levels at each stage, treating model card retrieval as a sliding window operation that balances comprehensive coverage with computational efficiency.

\subsection{Stage I: Iterative Reasoning and Dual-Stream Retrieval}

Stage I operationalizes a closed-loop reasoning-retrieval cycle designed to navigate the high-cardinality model space. Rather than treating retrieval as a static, one-shot operation, HuggingR$^4$ employs an agentic paradigm that autonomously refines search parameters based on intermediate feedback, progressively distilling the global repository into a high-precision candidate set.
\paragraph{Semantic Distillation of Model Cards.}
To maximize the signal-to-noise ratio within the prompt context, we perform semantic distillation on the raw model cards. For any $m_i \in \mathcal{D}$, we apply a transformation $\Psi: C(m_i) \to \tilde{C}(m_i)$ that prunes non-informative artifacts such as raw URLs, redundant citations, and organizational boilerplate, while canonicalizing technical specifications. This distillation ensures that the reasoning operates on high-density semantic representations, effectively mitigating prompt bloat and reducing token overhead by focusing on descriptors.
\paragraph{Intent Decomposition and Query Synthesis.}
% At each iteration $i$, the reasoning component takes as input the user query $q$, the Reasoning history $H_{i-1}$, the task category $c$ and generates a strategic retrieval query $s_i$. The reasoning process can be formalized as:
% \begin{equation}
%     s_i = LLM_{reason}(Instruct, q, H_{i-1}, c)
% \end{equation}
% where $LLM_{reason}$ represents the reasoning function parameterized. This function performs several key operations: 1) decomposing the user query into specific task requirements, domain constraints, and performance expectations; 2) generating targeted search terms that maximize the likelihood of retrieving relevant models while minimizing noise; and 3) choosing between Direct Retrieval and Metadata Retrieval based on the query characteristics.
At each iteration $t$, a reasoning agent $\mathcal{F}_{\text{reason}}$ synthesize a strategic search descriptor $s_t$ based on the user query $q$, the task category $c$, and the historical reasoning trace $H_{t-1}$. This process is formalized as:
\begin{equation}
s_t = \mathcal{F}_{\text{reason}}(q, H_{t-1}, c).
\end{equation}
The agent performs three critical operations:

\begin{itemize}
    \item \textbf{Functional decomposition:} translating high-level user needs into discrete technical constraints (\textit{e.g.}, ``low-latency'', ``quantized'', ``domain-specific'').
    \item \textbf{Lexical mapping:} generating targeted search terms that bridge the gap between user vernacular and the technical terminology prevalent in model repository metadata.
    \item \textbf{Policy selection:} adaptively deciding between direct semantic retrieval and metadata constrained retrieval based on the specificity of the intent.
\end{itemize}

\paragraph{Dual-Stream Adaptive Retrieval.}

To account for the inherent sparsity of repository metadata, we implement a \textit{dual-stream retrieval} mechanism. Depending on the agent's assessment, the system selects the appropriate embeddings:
\begin{itemize}
    \item \textbf{Direct stream}: encoding the full distilled description $\tilde{\mathcal{T}}_{\text{desc}}(m_i)$ into a dense vector $\mathbf{v}_{m_i}^{\text{full}} \in \mathbb{R}^d$, capturing broad functional nuances.
    \item \textbf{Metadata stream}: employing the structured attributes $\mathcal{A}_{\text{struct}}(m_i)$ to generate a constrained vector $\mathbf{v}_{m_i}^{\text{meta}} \in \mathbb{R}^d$, prioritizing hard-constraint satisfaction (\textit{e.g.}, language, dataset or license).
\end{itemize}
We employ \textit{multi-query augmentation} to enhance retrieval robustness. The system generates a set of $K$ semantically diverse variants $\mathcal{S}_t = \{ s_t^{(1)}, \dots, s_t^{(K)} \}$, which are concatenated into a composite query $s_t^{*} = \bigoplus_{i=1}^K s_t^{(i)}$. The final retrieval set $\mathcal{M}_{\text{reason}}$ is identified via cosine similarity:

\begin{equation}
\mathcal{M}_{\text{reason}} = \left\{ m_i \in \mathcal{D} \mid \text{sim}(s_t^{*}, \mathbf{v}_{m_i}^{t}) \in \text{Top-}k \right\},
\end{equation}
where $\text{sim}(\cdot)$ denotes the normalized inner product, and $\mathbf{v}_{m_i}^{t}$ is the embedding from the selected stream. The process terminates when $|\mathcal{M}_{\text{reason}}| \leq N$ (where $N$ is a predefined hyperparameter), ensuring the subsequent refinement stage remains computationally tractable.

% To ensure retrieval quality, we implement a Failure Tracing Mechanism that addresses the metadata incompleteness issue prevalent in Hugging Face model cards. When Metadata Retrieval is employed, the system performs a quality check by comparing the results with a corresponding Direct Retrieval operation. If the semantic similarity between the two result sets falls below a predefined threshold $\theta$, indicating potential metadata-induced retrieval failure, the system discards the current results, and backtracks to the previous reasoning step.

\paragraph{Failure Tracing and Backtracking.}

Recognizing that metadata in public repositories is often incomplete or misleading, we introduce a failure tracing mechanism. When a metadata-constrained search is performed, the system validates the result set by comparing its semantic density with a parallel direct stream sample. If the similarity divergence exceeds a predefined threshold $\theta$, indicating a potential metadata-induced failure, the system triggers a backtracking event. This forces the reasoning agent to revise its query synthesis strategy, thereby preventing the propagation of retrieval errors into the refinement stage.

% As shown in Figure~\ref{fig:3}, under the sliding window strategy, the system can only access the IDs of the top-k candidate models (as shown in blue of Figure~\ref{fig:3}) at this step, without accessing the full model cards. To make effective use of this limited information, the step adopts a ``reasoning-retrieval'' architecture, which guides the LLMs to shift their reasoning focus from the model cards themselves to the user query and the retrieval results. This not only significantly reduces token consumption and increases the precision of model selection, but also presents the reasoning process intuitively to the user, thereby improving the interpretability of the system.

\subsection{Stage II: Fine-Grained Semantic Refinement}

While Stage I prunes the search space through coarse-grained vector similarity, Stage II facilitates fine-grained semantic refinement. At this stage, the framework transitions from identifying relevant models to identifying the optimal one. This stage is critical because embedding-based retrieval often fails to distinguish between models with similar high-level descriptions but divergent technical nuances, such as specific quantization formats, library dependencies, or edge-case performance characteristics.

\paragraph{High-Fidelity Candidate Adjudication.}

Once the candidate set $\mathcal{M}_{\text{reason}}$ is reduced to a cardinality $|\mathcal{M}_{\text{reason}}| \leq N$, the refinement agent $\mathcal{F}_{\text{refine}}$ is granted access to the complete model cards $\tilde{C}(\mathcal{M}_{\text{reason}})$. These real model card provide the LLM with the necessary context to perform multi-dimensional cross-verification.

The refinement process is formalized as a selection function:
\begin{equation}
\mathcal{M}_{\text{refine}} = \mathcal{F}_{\text{refine}}(q, \tilde{C}(\mathcal{M}_{\text{reason}})).
\end{equation}
The agent performs a semantic analysis, evaluating candidates against the user query $q$ across several axes:
\begin{itemize}
    \item \textbf{Technical compatibility:} ensuring the model's architecture and dependencies align with the user's infrastructure.
    \item \textbf{Performance benchmarking:} analyzing qualitative and quantitative metrics reported in the model cards to verify suitability for the target domain.
    \item \textbf{Instruction adherence:} confirming that the model's intended use-case matches the operational requirements.
\end{itemize}
\paragraph{Computational Tractability via the Sliding Window Mechanism.}

We refer to the aforementioned approach as a sliding window mechanism: across Stage I iterations, the system progressively slides its focus across the model space, maintaining a fixed-size window $|\mathcal{M}_{\text{reason}}| \leq N$ of candidates for Stage II analysis. Unlike text-based sliding windows, this operates over the model candidate space.

The refinement stage drives computational efficiency by restricting high-fidelity LLM analysis to $\mathcal{M}_{\text{reason}}$, capping context consumption at $O(N \cdot L)$ where $L$ is the average model card token length. Since $N$ is much smaller than the full repository size $|\mathcal{D}|$, this bypasses the $O(|\mathcal{D}| \cdot L)$ complexity of direct Prompting and avoids ``lost-in-the-middle'' effects in long-context processing. Crucially, the approach decouples computational cost from repository size, enabling our approach to scale to millions of models.

\subsection{Stage III: Meta-Cognitive Reflection}

The final stage of HuggingR$^4$ introduces a self-correction Reflection layer, acting as a meta-cognitive supervisor. This stage is designed to mitigate the risks of model hallucination and false positive selections that may occur when the reasoning engine operates on partial information during the initial retrieval phase.

% \paragraph{High-Fidelity Candidate Adjudication.}

Due to the architectural decoupling in Stage I, where the system primarily reasons over model identifiers and distilled metadata, the agent may occasionally optimize for a ``local maximum" that appears suitable but lacks specific functional requirements found only in the deep documentation.
To address this, $\mathcal{F}_{\text{reflect}}$ executes a ``zero-trust'' audit. Given the refined candidate $\mathcal{M}_{\text{refine}}$ and the original query $q$, the reflection agent performs a rigorous verification:

\begin{equation}
\mathcal{M}^* = \mathcal{F}_{\text{reflect}} (q, \tilde{C}(\mathcal{M}_{\text{refine}})).
\end{equation}

The agent is tasked with identifying any misalignment between the model’s actual capabilities (e.g., specific tensor constraints, licensing restrictions, or hardware requirements) and the user's intent. If a discrepancy is detected, the agent documenting why $\mathcal{M}_{\text{refine}}$ is insufficient.

\paragraph{Dynamic Sliding Window and State Transition}

% As illustrated in Figure~\ref{fig:3}, we propose a novel sliding window strategy to dynamically manage the system’s access to candidate models throughout a multi-stage decision process. Each candidate model is treated as a window, and different levels of access are granted at different stages. This strategy effectively controls token consumption while maintaining selection accuracy. Specifically:
% a) In Step 1, the system only accesses the model IDs of the Top-k candidates (blue windows), significantly reducing token usage; b) In Step 2, when the number of candidate models falls below N, the system is granted access to the full model cards of these candidates (red windows). This enables the LLM to make detailed comparisons; c) In Step 3, the system runs a reflection on the selected model (dark yellow window) to determine whether it can satisfy the user request. If it fails, the red and yellow windows are frozen, and the blue window slides right by N positions.

% \subsection{Dynamic Sliding Window and State Transition}

The core innovation of our reflection process lies in its integration with the Sliding Window Strategy. Rather than terminating upon a failed validation, the framework initiates a state-space transition:
\begin{enumerate}
    \item \textbf{Freezing:} The current window of candidates $\mathcal{M}_{\text{reason}}$ is marked as ``exhausted" and cached in the reasoning history $H_t$ to prevent redundant processing.
    \item \textbf{Sliding:} The selection window $\mathcal{W}_t$ slides forward by $N$ positions in the global ranked list:

    \begin{equation}
    \mathcal{W}_{t+1} \gets \text{Slide}(\mathcal{W}_t, N).
    \end{equation}

    \item \textbf{Recursive Recovery:} The system re-enters Stage I. However, the reasoning agent $\mathcal{F}_{reason}$ now operates with the rich context of previous failure traces, adjusting its search parameters and exploring the next partition of the model space under improved heuristic guidance.
\end{enumerate}
This recursive mechanism ensures that HuggingR$^4$ maintains a high degree of recall robustness. By treating the repository as a series of evaluation windows, the framework can recover from initial retrieval biases without incurring the high tagging costs associated with global re-analysis, thus effectively striking the optimal balance between exhaustive search and computational simplicity.

% \begin{figure}[htb]
% \centering
% \includegraphics[width=1.0\columnwidth]{xin3-2.png}

% \caption{T}
% \label{fig1}
% \end{figure}

% \begin{figure}[htb]
% \centering
% \includegraphics[width=1.0\columnwidth]{xin3-3.png}

% \caption{T}
% \label{fig1}
% \end{figure}

\section{Experiments}

\begin{table*}[t]
\caption{Performance comparison of different LLMs on the single-task dataset. We evaluate HuggingGPT (baseline) and our HuggingR$^4$ across proprietary and open-source models. ``--'': API unsupported; HuggingR$^{4*}$: the retrieval-only version without refinement.}
\label{tab:1}
\vspace{-8pt}
\begin{center}
% \small
% \setlength{\tabcolsep}{2mm}
\begin{tabular}{l cc cc cc}
\toprule
& \multicolumn{2}{c}{\textbf{HuggingGPT \cite{shen2023hugginggpt}}} & \multicolumn{2}{c}{\textbf{HuggingR$^{4*}$}} & \multicolumn{2}{c}{\textbf{HuggingR$^{4}$}} \\
\cmidrule(lr){2-3} \cmidrule(lr){4-5} \cmidrule(lr){6-7}
\textbf{LLM Agent} & Workability & Reasonability & Workability & Reasonability & Workability & Reasonability \\
\midrule
GPT-4o-mini & 65.52 & 49.21 & 84.77 & 75.56 & \textbf{92.03} & 82.46 \\
GPT-4o & 75.20 & 60.70 & 84.15 & 75.16 & 91.14 & 82.09 \\
GPT-4.1-mini & 66.44 & 50.79 & 85.00 & 75.79 & 91.04 & 82.38 \\
GPT-4.1 & 74.80 & 60.73 & 85.27 & 75.72 & 91.43 & \textbf{83.86} \\
Qwen3-235b-a22b & 72.54 & 58.46 & 85.00 & 75.14 & 86.90 & 77.85 \\
Claude-Sonnet-4 & 78.64 & 64.86 & 86.02 & 76.19 & 90.85 & 81.59 \\
Deepseek-R1 & 78.94 & 66.83 & 86.35 & 77.79 & -- & -- \\
Gemini-2.5-Flash & 81.20 & 68.50 & 84.81 & 75.66 & -- & -- \\
Qwen2.5-7b & 56.59 & 41.83 & 82.78 & 73.82 & 85.73 & 76.31 \\
\bottomrule
\end{tabular}
\end{center}
\vspace{-10pt}
\end{table*}

\subsection{Dataset Construction}
To evaluate our approach, we constructed a comprehensive dataset comprising three components. The complete dataset construction process and associated prompts are detailed in the Appendix~\ref{sec:Data}.

% \subsubsection{Model Cards}
\noindent\textbf{Model Cards.} The dataset includes 37 different tasks from HuggingFace. For each task, we ranked models based on the number of likes and selected the top-ranking models, resulting in a candidate pool of 1,110 models. The complete task distribution is provided in Appendix Table~\ref{tab:task_distribution}.

\noindent\textbf{Single-Task User Requests.} We constructed 1,016 single-task requests covering diverse domains represented by the candidate models. Each request was annotated by domain experts across two evaluation dimensions: \textit{Workability} (whether the model satisfies basic user needs) and \textit{Reasonability} (whether the model aligns with domain-specific preferences). 

% The detailed annotation protocol and quantitative 
% criteria are described in Section~\ref{sec:metrics}.

% \noindent\textbf{Multi-Task User Requests.} We leveraged GPT-4o to combine the single-task requests into multi-task scenarios. These generated multi-task requests were reviewed and refined by domain experts to ensure their feasibility and coherence. A total of 13,383 multi-task user requests were curated for evaluation.
\noindent\textbf{Multi-Task User Requests.} 
We leveraged GPT-4o to combine single-task requests into multi-task scenarios, with each scenario containing 2-5 tasks. All 13,383 generated requests were manually reviewed by five domain experts, who verified: (1) \textit{domain consistency} (do all tasks share coherent domain preferences, e.g., all biomedical or all social media?), (2) \textit{semantic coherence} (do tasks form a meaningful workflow?), and (3) \textit{technical feasibility} (are required models available on HuggingFace?). Requests failing any criterion were either revised or discarded. After quality control, all 13,383 requests were retained as multi-task scenarios.

\vspace{-8pt}

\subsection{Experimental setup}
\vspace{-5pt}
% In our experiments, we thoroughly evaluated the model selection performance of HuggingR$^{4*}$ and HuggingR$4$ across 8 different LLMs. These models cover several popular series, including GPT, Claude, Qwen and DeepSeek, all of which are accessible via API calls. In the configuration for HuggingR$^{4*}$, the Multi-Query parameter is set to 4, and the embedding model chosen is text-embedding-3-large. Additionally, the LLM parameters are set to their default values, including temperature, Top-K, and Top-P, to ensure consistency and reproducibility. For the sliding window strategy, N and K are set to 3 and 5, respectively.
% In our experiments, we thoroughly evaluated the model selection performance of HuggingR$^{4*}$ and HuggingR$4$ across 8 different LLMs. These models cover several popular series, including GPT, Claude, Qwen and DeepSeek, all of which are accessible via API calls.

In our experiments, we thoroughly evaluated the model selection performance of HuggingR$^{4*}$ and HuggingR$^4$ across 9 different LLMs. The evaluated models span several popular series, including GPT, Claude, Qwen, and DeepSeek.

% The experimental configuration includes: Multi-Query number set to 4, text-embedding-3-large as the embedding model, sliding window parameters N=3 and K=5, and failure tracing threshold $\theta=80\%$. Complete parameter specifications are provided in the appendix.
The experimental configuration is as follows: Multi-Query number is set to 4, with text-embedding-3-large as the embedding model. For the sliding window strategy, parameters $N$ and $k$ are configured to 3 and 5, respectively. The failure tracing threshold $\theta$ is set to 80\%. To ensure experimental statistical significance, we report the average results across three independent runs for all evaluations. Complete parameter settings are provided in the Appendix~\ref{sec:setup}.

We employ HuggingGPT as our baseline, as it is a representative and the only query-driven model selection method, which directly embeds model cards into prompts for LLM-based selection. Following the official implementation, we apply a 100-character truncation limit to each model card.

Additionally, based on HuggingR$^4$, we introduce a pure retrieval variant, denoted as HuggingR$^{4*}$, which retains only the multi-stage retrieval mechanism and achieves maximal token savings by removing the refinement and Reflection.

% \vspace{-15pt}

\subsection{Evaluation Metrics}
\label{sec:metrics}
We propose two key evaluation metrics to assess model selection quality: \textit{Workability} and \textit{Reasonability}. These metrics capture different aspects of selection quality, from functional correctness to domain-specific appropriateness.
\vspace{-12pt}

\setlength{\leftmargini}{10pt}
\begin{itemize}
\setlength{\itemsep}{-0.3em}
    \item \textbf{Workability (Task Compatibility):} We define a model as \textit{workable} if it satisfies three binary criteria: (1) task type matches the request, (2) input/output formats are compatible, and (3) the model is executable on Hugging Face Hub or locally. A model receives a Workability score of 1 if all criteria are met, 0 otherwise. This yields an average of 8.3 workable models per request.
    
    \item \textbf{Reasonability (Domain Appropriateness):} Among workable models, we define a model as \textit{reasonable} if it satisfies two criteria: (1) trained on domain-relevant datasets (verified through model card metadata), and (2) achieves competitive performance (within top-5 reported metrics among candidate models). This yields an average of 2.1 reasonable models per request.
\end{itemize}
\vspace{-10pt}
% Five domain experts independently verified 200 randomly sampled annotations, achieving 96.5\% agreement (Cohen's $\kappa = 0.94$), confirming the reliability of our criteria.

% To validate the reliability of our annotation criteria, five domain experts independently annotated 200 randomly sampled requests, achieving 96.5\% agreement (Cohen's $\kappa = 0.94$). 

Detailed annotation guidelines and inter-annotator agreement analysis are provided in Appendix~\ref{sec:setup}.

\subsection{Results and Analysis}
Table~\ref{tab:1} presents the comprehensive evaluation results of HuggingR$^4$ against the baseline HuggingGPT across different LLM backbones. Our proposed method consistently outperforms the baseline across all tested models, demonstrating substantial improvements in both workability and reasonability metrics. Specifically, HuggingR$^4$ achieves the most balanced performance on GPT-4o-mini with 92.03\% workability and 82.46\% reasonability, representing improvements of 26.51\% and 33.25\% respectively over HuggingGPT. The retrieval-only variant HuggingR$^{4*}$ shows consistent performance across different LLMs since the retrieval process primarily relies on Multi-Query Generation rather than the underlying model's reasoning capabilities.

Notably, while stronger LLMs generally yield better baseline performance, our framework's effectiveness varies across different models due to their distinct reasoning patterns. Claude-Sonnet-4 exhibits relatively lower performance due to its tendency toward over-analysis during the reflection stage, leading to overly strict model filtering that eliminates suitable candidates. Moreover, Qwen3-235b-a22b shows suboptimal results due to excessive iterations in the initial reasoning-retrieval phase, struggling to converge efficiently. Additionally, our method demonstrates strong effectiveness even on smaller open-source models: with Qwen2.5-7B, HuggingR$^4$ achieves 85.73\% workability, representing a 29.14\% improvement over HuggingGPT, suggesting that our method can effectively enhance smaller models' performance. These observations highlight the importance of balancing thorough analysis with practical efficiency in progressive reasoning frameworks.

\begin{table}[t]
\centering
\setlength{\tabcolsep}{1.5mm}
\caption{Performance of HuggingR$^{4*}$ and HuggingR$^4$ with different embedding models on GPT-4o-mini.}
\vspace{-3pt}
\label{tab:embedding}
\begin{tabular}{lcc}
\toprule
\textbf{Method} & {Workability} & {Reasonability} \\
\midrule
\multicolumn{3}{l}{\textit{HuggingR$^{4*}$}} \\
\quad text-embedding-3-large & 84.77 & 74.56 \\
\quad jina-embeddings-v3     & 84.81 & 73.75 \\
\quad text-embedding-3-small & 83.04 & 70.93 \\
\midrule
\multicolumn{3}{l}{\textit{HuggingR$^4$}} \\
\quad text-embedding-3-large & \textbf{92.03} & \textbf{82.46} \\
\quad jina-embeddings-v3     & 90.62 & 80.67 \\
\quad text-embedding-3-small & 90.12 & 79.95 \\
\bottomrule
\end{tabular}
\vspace{-20pt}
\end{table}

We also evaluate the performance of our methods with different embedding models. As shown in Table~\ref{tab:embedding}, the performance of HuggingR$^4$ generally scales with the capacity of embedding models, while the consistently high performance across all models (90\%+ workability) demonstrates the robustness of our approach to different semantic encoders. Notably, even with the smaller text-embedding-3-small, HuggingR$^4$ achieves 90.12\% workability, only 1.91\% lower than the best configuration, demonstrating that our method provides flexibility in balancing computational efficiency and performance. This robustness enables practitioners to select embedding models based on their specific deployment constraints without significantly compromising recommendation quality.

We further evaluate HuggingR$^4$ in multi-task scenarios adopting the same task planing as HuggingGPT. As shown in Table~\ref{tab:2}, HuggingR$^4$ achieves workability and reasonability rates of 85.03\% and 75.73\% respectively, still substantially outperforming the original baseline methods. However, compared to single-task performance (92.03\% workability, 82.46\% reasonability), we observe a notable decline in both metrics. This performance gap primarily stems from the increased complexity of task planning and decomposition required for multi-task scenarios, indicating that our current approach for handling task orchestration and dependency management requires further optimization.

\begin{table}[t]
\caption{Performance comparison on the multi-task dataset using GPT-4o-mini with text-embedding-3-large.}
\label{tab:2}
\vspace{-5pt}
\begin{center}
% \small
\begin{tabular}{l cc}
        \toprule
        {\textbf{Method}} & Workability & Reasonability\\
        \midrule
        HuggingGPT & 68.08 & 51.44  \\
        HuggingR$^{4*}$ & 78.27 & 69.86\\
        HuggingR$^{4}$ & \textbf{85.03} & \textbf{75.73} \\
        
        \bottomrule
       
\end{tabular}
\end{center}
% \vspace{-2mm}
\vspace{-10pt}

\end{table}

\begin{table}[t]
\centering
\caption{Comparison of HuggingR$^4$ with Other Paradigms}
\vspace{-5pt}
\label{tab:comparison}
\small
\setlength{\tabcolsep}{1.0mm}
\begin{tabular}{l|cccc}
\toprule
\textbf{Method} & 
\textbf{Input Type} & 
\textbf{\# Tasks} & 
\textbf{\# Models} & 
\textbf{Online} \\
\midrule
AutoMRM & Dataset & 1 & 11 & \xmark \\
TransferGraph & Dataset & 2 & 348 & \xmark \\
ModsNet & Dataset & 3 & 3,194 & \xmark \\
HuggingGPT & User Query & 24 & \textgreater 10K & \cmark \\
\rowcolor{blue!10}
\textbf{HuggingR$^4$} & \textbf{User Query} & \textbf{37} & \textbf{\textgreater 10K} & \cmark \\
\bottomrule
\end{tabular}
\vspace{-15pt}
\end{table}

\subsection{Comparison with Other Paradigms}
% While Section~\ref{sec:results} presents quantitative comparisons, we now clarify why direct comparison with other model recommendation systems is infeasible due to fundamental paradigm differences.
Table~\ref{tab:comparison} reveals two distinct paradigms with fundamentally different design philosophies. Data-driven methods (AutoMRM~\cite{li2023automrm}, TransferGraph~\cite{li2024model}, ModsNet~\cite{wang2023selecting}) require \textit{(1) expert knowledge} to extract dataset meta-features (e.g., statistical properties, embeddings), \textit{(2) offline training} on pre-collected model-task pairs, and \textit{(3) limited task coverage} (1--3 specific tasks per system). In contrast, our query-driven approach offers three key advantages: \textbf{(1) User-friendly.} Non-experts can express requirements in natural language (e.g., \textit{"I need sentiment analysis for Chinese movie reviews"}); \textbf{(2) Online adaptability.} Seamlessly integrates with evolving model communities (Hugging Face's daily uploads) and can be deployed as an agent; \textbf{(3) Broad coverage.} Handles 37 diverse tasks spanning NLP, vision, audio, and multimodal domains with 10K+ models. 

\section{Ablation Study}
% To comprehensively evaluate the effectiveness of our system design, we conduct ablation experiments to analyze the roles of key modules, the impact of critical hyperparameters, and the token consumption.

\subsection{Impact of Individual Modules}
In order to assess the contribution of each individual module, we conduct ablation experiments by removing specific components. Table~\ref{tab:3} reports the resulting performance changes.

For HuggingR$^{4*}$, removing the Semantic Distillation results in a significant decrease in both Workability (-3.41) and Reasonability (-4.30). This is because semantic distillation substantially enhances model card embedding quality, increasing the proportion of semantically meaningful vectors and thereby improving retrieval effectiveness. Disabling Metadata stream retrieval also leads to substantial performance degradation (-5.80 in Workability and -5.74 in Reasonability), highlighting the benefits of using metadata labels to filter relevant tasks and narrow the search space.

% Conversely, deactivating Metadata-Driven Dataset Retrieval causes negligible performance loss. This is primarily because most model cards on HuggingFace contain incomplete dataset metadata, thereby diminishing the utility of this signal given the present state of metadata availability.

For HuggingR$^4$, deactivating the Failure Tracing and Backtracking results in the most substantial performance degradation (-4.80 Reasonability). This component is critical to addressing missing or incomplete metadata in relevant models. Its absence increases the likelihood of the system returning incorrect candidates stemming from undetected metadata mismatches. The removal of either the Self-reflection or the Sliding Window Strategy further impairs performance. Both components empower the system to explore beyond the initially retrieved top-ranked candidates, expanding the candidate pool.
Critically, these modules facilitate multi-round reasoning, enabling iterative refinement of the retrieval and selection processes to enhance robustness and accuracy.

\begin{table}[t]

\begin{center}
\caption{
Ablation study on individual module contributions using GPT-4o-mini with text-embedding-3-large.
}
\label{tab:3}
% \small
\setlength{\tabcolsep}{0.8mm}
% \centering
\begin{tabular}{l cc}
        \toprule
        {\textbf{Method}} & Workability & Reasonability\\
        \midrule
         
        \textit{HuggingR$^{4*}$} & \textbf{84.77} & \textbf{75.56}  \\
        w/o Multi-Query & 82.94 & 71.63\\
        w/o Semantic Distillation & 81.36 & 71.26\\
        w/o Metadata Stream & 78.97 & 69.82\\
        % w/o Dataset Retrieval & 84.38 & 74.83\\
        \midrule
        \textit{HuggingR$^4$} & \textbf{92.03} & \textbf{82.46}  \\
        w/o Failure Tracing & 88.09 & 77.66\\
        w/o Self-reflection & 85.33 & 80.21\\
        w/o Sliding Window Strategy & 90.98 & 81.30\\
        
        \bottomrule
       
\end{tabular}
\end{center}
% \vspace{-2mm}
\vspace{-10pt}

\end{table}

\begin{table}[t]
\caption{Performance impact of hyperparameters $N$ (window size) and $k$ (number of retrieved candidate models) in the sliding window strategy using GPT-4o-mini with text-embedding-3-large.}
\centering
\small
\setlength{\tabcolsep}{1mm}
\begin{tabular}{ccc|ccc}
\toprule
\multicolumn{3}{c|}{\textbf{$N$ (k=5)}} & \multicolumn{3}{c}{\textbf{$k$ (N=3)}} \\
$N$ & Workability & Reasonability & $k$ & Workability & Reasonability \\
\midrule
1 & 88.98 & 79.82 & 3 & 90.98 & 81.30 \\
2 & 91.54 & 81.99 & 4 & 91.73 & 82.14 \\
3 & 92.03 & 82.46 & 5 & \textbf{92.03} & \textbf{82.46} \\
4 & \textbf{92.35} & \textbf{83.07} & 6 & 91.63 & 81.79 \\
5 & 92.32 & 82.48 & 7 & 91.34 & 81.99 \\
\bottomrule
\end{tabular}

\label{tab:4}
\vspace{-15pt}
\end{table}

\vspace{-5pt}
\subsection{Analysis of the Sliding Window Strategy}
We further investigate the impact of the two key hyperparameters in the sliding window strategy: $k$ and $N$. As shown in Table~\ref{tab:4}, increasing $N$ consistently improves both Workability and Reasonability scores, with peak performance at $N{=}4$. However, larger $N$ values also lead to higher token consumption due to more full model cards being processed in stage II. To strike a balance between performance and efficiency, we select $N{=}3$ as the default configuration. In contrast, varying $k$ shows a relatively stable trend, where performance improves up to $k{=}5$ but slightly degrades when $k$ exceeds 5. This indicates that an appropriate number of initial candidates is beneficial, but overly large candidate pools may introduce noise and reduce effectiveness.

% \subsection{Impact of Failure Tracing}

% To evaluate the effectiveness of the failure tracing mechanism, we conduct ablation studies by varying the similarity threshold $\theta$ that triggers the traceback process. Table~\ref{tab:ablation_theta} presents the results across different threshold settings. We observe a consistent improvement trend as $\theta$ increases from 0\% to 80\%, with performance peaking at $\theta=80\%$ (92.03\% workability and 82.46\% reasonability). Disabling the mechanism ($\theta=0\%$) yields the lowest performance (88.09\%/77.66\%), while an overly strict threshold ($\theta=100\%$) leads to slight degradation (91.24\%/81.99\%), indicating that moderate thresholds best balance failure detection and false positive avoidance.

% \begin{table}[t]
% \centering
% \small
% \caption{Ablation study on the failure tracing threshold $\theta$ using HuggingR$^4$ with GPT-4o-mini and text-embedding-3-large.}
% \label{tab:ablation_theta}
% \begin{tabular}{c|cc}
% \toprule
% \textbf{Threshold} $\theta$ & \textbf{Workability (\%)} & \textbf{Reasonability (\%)} \\
% \midrule
% 0\%      & 88.09 & 77.66 \\
% 40\%               & 90.45 & 80.41 \\
% 60\%               & 91.73 & 81.69 \\
% \textbf{80\%} & \textbf{92.03} & \textbf{82.46} \\
% 100\%              & 91.24 & 81.99 \\
% \bottomrule
% \end{tabular}
% \vspace{-13pt}
% \end{table}

\vspace{-5pt}
\subsection{Analysis of Token Usage}

\begin{figure}[t]
\centering
\centerline{\includegraphics[width=1.0\columnwidth]{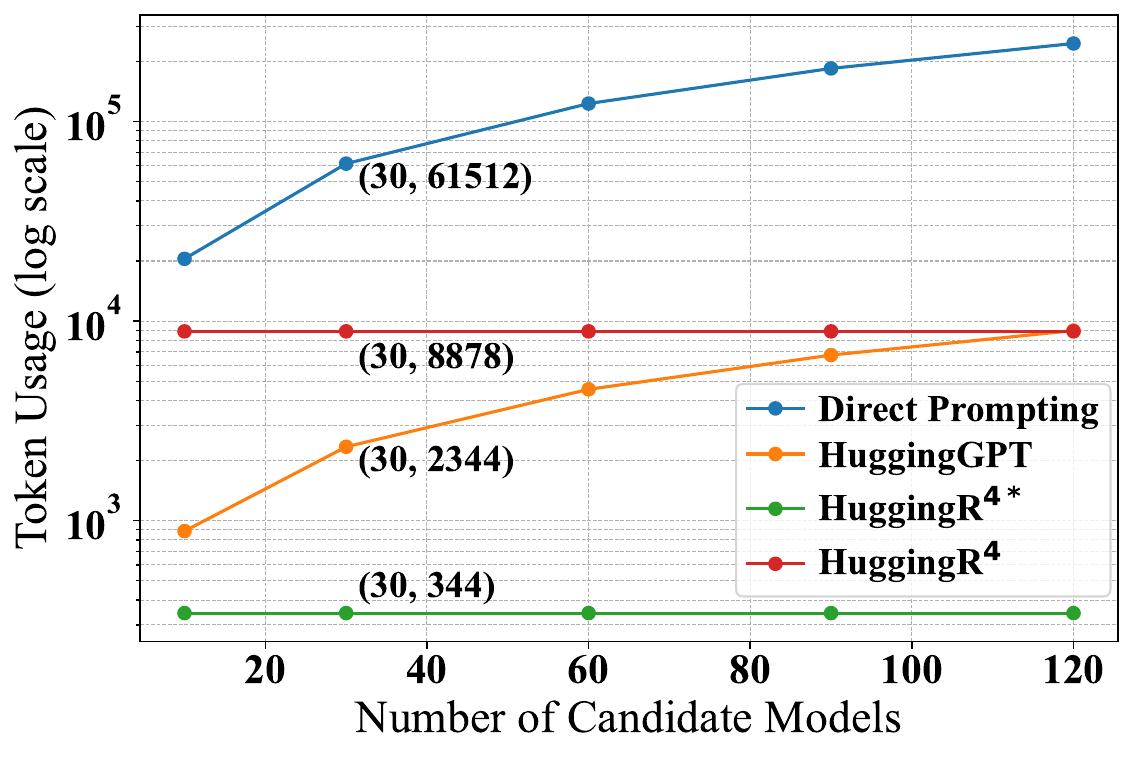}}
\vspace{-8pt}
\caption{Token usage comparison across different numbers of candidate models using GPT-4o-mini with text-embedding-3-large (log scale). Our HuggingR$^4$ and HuggingR$^{4*}$ maintain constant token consumption through the sliding window strategy, while Direct Prompting and HuggingGPT scale linearly with the number of candidates. At 30 candidates, HuggingR$^4$ achieves 85.6\% reduction compared to Direct Prompting.}
\label{fig:4}
\vspace{-10pt}
\end{figure}

In practical applications, token consumption is a key efficiency indicator. A major advantage of our approach is that it avoids increasing token usage as the number of candidate models grows, making it highly scalable. To illustrate this, we compare four representative methods with varying sizes of the candidate pool.
As shown in Figure~\ref{fig:4}, both Direct Prompting and HuggingGPT consume significantly more tokens when handling more models, due to their reliance on directly embedding large model cards into the prompt. In contrast, HuggingR$^4$'s token consumption is independent of the number of candidate models, as they operate on retrieval results. Notably, HuggingR$^{4*}$ achieves high model selection performance while consuming only a minimal amount of tokens. Due to its multi-step reasoning process, HuggingR$^4$ requires slightly more tokens. However, it achieves superior performance when handling a larger number of candidate models. Therefore, our proposed methods not only excel in model selection capability but also provide significant advantages in token consumption.

\section{Conclusion}

In this paper, we introduce HuggingR$^4$, a progressive reasoning framework that addresses the critical challenge of selecting optimal AI model companions from large-scale repositories like HuggingFace. To evaluate our approach, we created the first forward-labeled user request dataset and conducted extensive experiments along two key dimensions: workability and reasonability. Results show that our approach better aligns with users' fine-grained preferences and maintains stable token consumption, regardless of the number of candidate models, leading to significant reductions in overall token usage and improved efficiency.

\section*{Impact Statement}

This paper presents work whose goal is to advance the field of Machine Learning. There are many potential societal consequences of our work, none which we feel must be specifically highlighted here.

\nocite{langley00}

\bibliography{example_paper}
\bibliographystyle{icml2026}

%%%%%%%%%%%%%%%%%%%%%%%%%%%%%%%%%%%%%%%%%%%%%%%%%%%%%%%%%%%%%%%%%%%%%%%%%%%%%%%
%%%%%%%%%%%%%%%%%%%%%%%%%%%%%%%%%%%%%%%%%%%%%%%%%%%%%%%%%%%%%%%%%%%%%%%%%%%%%%%
% APPENDIX
%%%%%%%%%%%%%%%%%%%%%%%%%%%%%%%%%%%%%%%%%%%%%%%%%%%%%%%%%%%%%%%%%%%%%%%%%%%%%%%
%%%%%%%%%%%%%%%%%%%%%%%%%%%%%%%%%%%%%%%%%%%%%%%%%%%%%%%%%%%%%%%%%%%%%%%%%%%%%%%
\newpage
\clearpage
\appendix
\onecolumn

%%%%%%%%%%%%%%%%%%%%%%%%%%%%%%%%%%%%%%%%%%%%%%%%%%%%%%%%%%%%%%%%%%%%%%%%%%%%%%%
\section{Methodology Implementation Details}
\label{sec:rationale}
\subsection{Methodology Prompt}

Our framework structures the LLM as a multi-stage reasoning assistant. Given user query $q$ and optional task label, the system prompt orchestrates:

\textbf{Stage I: Iterative Retrieval.} The system generates targeted search queries $s_t$ and applies filtering tools across multiple rounds, adaptively refining the strategy (broadening if too few results, narrowing if too many) until obtaining $|\mathcal{M}_{\text{reason}}| \leq N$ candidates. The prompt explicitly prohibits inferring capabilities from model names, enforcing metadata validation.

\textbf{Stage II: Comparative Refinement.} With $|\mathcal{M}_{\text{reason}}| \leq N$ candidates, a refinement tool performs deep comparison across: (1) task performance, (2) efficiency, (3) compatibility, and (4) community adoption. The LLM selects the optimal model $m^*$ through pairwise analysis with detailed justification.

\textbf{Stage III: Meta-Cognitive Reflection.} A reflection agent adversarially audits $\mathcal{M}_{\text{reflect}}$ against all requirements (language, dataset, size, license, hardware, special constraints). If verified, the system returns $\boxed{\texttt{MODEL\_NAME}}$; if any check fails, it outputs $\boxed{\texttt{UNCERTAIN}}$ with failure trace, triggering fallback Stage I with augmented query. The prompt as showcased below:

\begin{tcolorbox}
[enhanced, colback=gray!4, colframe=gray!100, sharp corners, 
leftrule={0pt}, rightrule={0pt}, toprule={0pt}, bottomrule={0pt}, 
left={7pt}, right={2pt}, top={3pt}, bottom={3pt},
overlay={\draw[gray!100, line width=2.1pt] ([xshift=1.0pt]frame.north west) -- ([xshift=1.0pt]frame.south west);
         \draw[gray!70, line width=2pt] ([xshift=3pt]frame.north west) -- ([xshift=3pt]frame.south west);}, frame hidden]

{You are an expert reasoning assistant for selecting models on HuggingFace. Your task is to \textbf{progressive reasoning}, iteratively filtering out the most appropriate model based on the user's input: \{user query\}.}

\textbf{{Stage I: Iterative Reasoning and Dual-Stream Retrieval}}
    
% \begin{itemize}
- Carefully analyze the user's requirements and generate targeted search queries.
- Use available filtering tools iteratively (e.g., filter by task, language, size, license).

- Refine your search strategy based on intermediate results, if you get too many results, add more constraints; if you get zero results, broaden your search.

- Continue retrieval cycles until you obtain {$N$} or fewer candidate models.

- Critical: Do NOT assume model capabilities based solely on model names. Always validate against model metadata.
% \end{itemize}

\textbf{{Stage II: Fine-Grained Semantic Refinement}}

- {Once you have $\leq \{N\}$ candidate models, use refinement tool to provide a selection of the one best model.}

- Analyze each candidate across multiple dimensions: Task alignment and performance benchmarks. Efficiency (inference speed, memory usage). Compatibility (frameworks, hardware). Community adoption and maintenance status.

- Select the best model based on this comprehensive analysis.

\textbf{{Stage III: Meta-Cognitive Reflection}}

- {Verify the selected model satisfies all user criteria: language, dataset compatibility, model size, type, and special requirements}

- {If the model meets all these criteria, and you have information to recommend this model, return the final result: $\boxed{\texttt{MODEL\_NAME}}$.}

- {If any of the criteria are not satisfied, system output: $\boxed{\texttt{UNCERTAIN}}$.}

\end{tcolorbox}

\subsection{Iterative Reasoning and Dual-Stream Retrieval Prompts}

% \subsubsection{Model Card Preprocessing} 
\paragraph{Semantic Distillation of Model Cards.}

Given that raw model cards often contain verbose documentation, redundant metadata, and heterogeneous formatting that can hinder retrieval efficiency, we perform semantic distillation on all model cards before framework execution. Specifically, we apply a transformation $\Psi: C(m_i) \to \tilde{C}(m_i)$ that removes non-informative artifacts (e.g., raw URLs, redundant citations, organizational boilerplate) while canonicalizing technical specifications and extracting essential fields. This preprocessing step maximizes the signal-to-noise ratio by retaining only the core information needed for model selection, such as model purpose, capabilities, and key technical descriptors, while eliminating elements that contribute to prompt bloat without aiding the reasoning process:

\begin{tcolorbox}
[enhanced, colback=gray!4, colframe=gray!100, sharp corners, 
leftrule={0pt}, rightrule={0pt}, toprule={0pt}, bottomrule={0pt}, 
left={7pt}, right={2pt}, top={3pt}, bottom={3pt},
overlay={\draw[gray!100, line width=2.1pt] ([xshift=1.0pt]frame.north west) -- ([xshift=1.0pt]frame.south west);
         \draw[gray!70, line width=2pt] ([xshift=3pt]frame.north west) -- ([xshift=3pt]frame.south west);}, frame hidden]

\textbf{Prompt:}
{You are a helpful assistant specialized in simplifying HuggingFace model cards. Your task is to extract only the fields relevant to model selection: id, downloads, likes, pipeline\_tag, task, meta, language, datasets, and description. For the description field, simplify the content by keeping only the most essential information that helps users understand the model's purpose and use case. Output only a clean, valid JSON object with the selected and simplified fields.}

\end{tcolorbox}

% \subsubsection{Retrieval Tool Prompt}
% \paragraph{Dual-Stream Retrieval Prompt.}
\paragraph{Dual-Stream Adaptive Retrieval Prompt.}

To enable flexible and precise model filtering from the large-scale repository while accounting for the inherent sparsity of repository metadata, we instruct the LLM to utilize a dual-stream retrieval mechanism in the main prompt. The system autonomously selects between two retrieval streams based on query analysis: direct stream and metadata stream. The LLM autonomously selects and combines these tools based on query analysis, enabling adaptive retrieval strategies.

The retrieval prompts are structured as follows:

\begin{tcolorbox}
[enhanced, colback=gray!4, colframe=gray!100, sharp corners, 
leftrule={0pt}, rightrule={0pt}, toprule={0pt}, bottomrule={0pt}, 
left={7pt}, right={2pt}, top={3pt}, bottom={3pt},
overlay={\draw[gray!100, line width=2.1pt] ([xshift=1.0pt]frame.north west) -- ([xshift=1.0pt]frame.south west);
         \draw[gray!70, line width=2pt] ([xshift=3pt]frame.north west) -- ([xshift=3pt]frame.south west);}, frame hidden]

\textbf{Tool 1: Direct Similarity Retrieval}

- {Used to find models similar to the user’s query.}

\textbf{Input:} {The retrieval query:}

\begin{quote}
    \texttt{<|begin\_similarity\_query|>
    Retrieval Query 
    <|end\_similarity\_query|>}
\end{quote}

\textbf{Output:} {The top-$k$ most relevant models:}

\begin{quote}
    \texttt{<|begin\_similarity\_result|>
    model 1, model 2 ... <|end\_similarity\_result|>}
\end{quote}

\end{tcolorbox}

\begin{tcolorbox}
[enhanced, colback=gray!4, colframe=gray!100, sharp corners, 
leftrule={0pt}, rightrule={0pt}, toprule={0pt}, bottomrule={0pt}, 
left={7pt}, right={2pt}, top={3pt}, bottom={3pt},
overlay={\draw[gray!100, line width=2.1pt] ([xshift=1.0pt]frame.north west) -- ([xshift=1.0pt]frame.south west);
         \draw[gray!70, line width=2pt] ([xshift=3pt]frame.north west) -- ([xshift=3pt]frame.south west);}, frame hidden]

\textbf{Tool 2.1: Metadata Language Retrieval}

- {Filters models based on the specified language.}

\textbf{Input:} {The ISO language code of user query:}
\begin{quote}
    \texttt{<|begin\_language\_query|>
    ISO language code
    <|end\_language\_query|>}
\end{quote}

\textbf{Output:} {The Top-k models:}
\begin{quote}
    \texttt{<|begin\_language\_result|>
    model 1, model 2 ... 
    <|end\_language\_result|>}
\end{quote}

\textbf{Tool 2.1: Metadata Dataset Retrieval}

- \textit{Filters models based on the dataset requirements.}

\textbf{Input:} {Describe the required dataset:}
\begin{quote}
    \texttt{<|begin\_dataset\_query|>
    Describe the required dataset\
    <|end\_dataset\_query|>}
\end{quote}

\textbf{Output:} {The Top-k models:}
\begin{quote}
    \texttt{<|begin\_dataset\_result|>
    model 1, model 2 ... 
    <|end\_dataset\_result|>}
\end{quote}

\textbf{Special Case:} {If some model cards are missing dataset labels, and no similarity search results are returned for the dataset retrieval, mark this dataset retrieval as untrustworthy.}

\end{tcolorbox}

% \subsubsection{Multi-Query Generation}
\paragraph{Multi-Query Augmentation.}

To enhance query diversity and improve retrieval coverage, we generate multiple search queries from different perspectives before performing the model retrieval. This multi-query approach allows us to capture various aspects of the user's intent and increases the likelihood of finding relevant models through different semantic angles. In our implementation, we set $K=4$ and use the following prompt to transform the user query into four distinct search queries:

\begin{tcolorbox}
[enhanced, colback=gray!4, colframe=gray!100, sharp corners, 
leftrule={0pt}, rightrule={0pt}, toprule={0pt}, bottomrule={0pt}, 
left={7pt}, right={2pt}, top={3pt}, bottom={3pt},
overlay={\draw[gray!100, line width=2.1pt] ([xshift=1.0pt]frame.north west) -- ([xshift=1.0pt]frame.south west);
         \draw[gray!70, line width=2pt] ([xshift=3pt]frame.north west) -- ([xshift=3pt]frame.south west);}, frame hidden]

\textbf{Prompt:}
{You are a helpful assistant in generating multiple search queries based on a single query entered by a user. This will allow us to use the user query to more accurately find the most similar model via Hugging Face model card embeddings. The generated queries should characterize the model itself (e.g., inputs to the model, uses of the model, datasets, kinds of languages, etc.) as needed by the user. You only need to output these {$K$} queries, each on a new line, with no other information.}
    
{Generate multiple search queries related to: \{question\}}
    
\textbf{Output:} {(4 queries)}

\end{tcolorbox}

The prompt instructs the LLM to focus on different model characteristics such as input types, use cases, training datasets, and supported languages. The output consists of four queries, each presented on a separate line without additional formatting or explanations. These diversified queries are then used in parallel during the reasoning-retrieval phase to maximize the breadth of candidate model discovery.

\subsection{Fine-Grained Semantic Refinement Prompts}

% \subsubsection{Refinement Tool Prompt} 
\paragraph{Refinement Tool Prompt.}

In the refinement step, the system obtains complete model cards for the candidate models identified in the coarse-grained retrieval phase. This enables more informed model selection by providing comprehensive descriptions rather than the simplified representations used in the initial retrieval.

The refinement process utilizes a specialized tool that fetches detailed model information from the vector database. This tool takes the candidate model list from Stage I and returns their complete model cards, allowing the LLM to make more nuanced decisions based on full specifications, capabilities, and usage guidelines:

\begin{tcolorbox}
[enhanced, colback=gray!4, colframe=gray!100, sharp corners, 
leftrule={0pt}, rightrule={0pt}, toprule={0pt}, bottomrule={0pt}, 
left={7pt}, right={2pt}, top={3pt}, bottom={3pt},
overlay={\draw[gray!100, line width=2.1pt] ([xshift=1.0pt]frame.north west) -- ([xshift=1.0pt]frame.south west);
         \draw[gray!70, line width=2pt] ([xshift=3pt]frame.north west) -- ([xshift=3pt]frame.south west);}, frame hidden]

\textbf{Tool 4: Get a detailed description of the model:}

- {Used to get the full model cards.}

\textbf{Input:} {Output model list of stage I:}
\begin{quote}
            \texttt{<|begin\_descriptions\_query|>
            model 1, model 2, model 3 \\
            <|end\_descriptions\_query|>}
\end{quote}

\textbf{Output:} {Complete description of these models:}

\begin{quote}
            \texttt{<|begin\_descriptions\_result|> \\
            \{description 1\}, \{description 2\}, 
            \{description 3\} \\
            <|end\_descriptions\_result|>}
\end{quote}

\end{tcolorbox}

% This tool design ensures that detailed model information is only retrieved for relevant candidates, maintaining efficiency while providing the comprehensive context necessary for accurate model selection.

\subsection{Meta-Cognitive Reflection Prompts}

% \subsubsection{Reflection Failure Handling}
\paragraph{Reflection Failure Handling.}

When the reflection module determines that the models selected in Stage II are inadequate for the user's request, the main prompt outputs $\boxed{\texttt{UNCERTAIN}}$, triggering a backtrack to Step 1 with an expanded search scope. The system employs the following prompt to guide the LLM through this recovery process:

\begin{tcolorbox}
[enhanced, colback=gray!4, colframe=gray!100, sharp corners, 
leftrule={0pt}, rightrule={0pt}, toprule={0pt}, bottomrule={0pt}, 
left={7pt}, right={2pt}, top={3pt}, bottom={3pt},
overlay={\draw[gray!100, line width=2.1pt] ([xshift=1.0pt]frame.north west) -- ([xshift=1.0pt]frame.south west);
         \draw[gray!70, line width=2pt] ([xshift=3pt]frame.north west) -- ([xshift=3pt]frame.south west);}, frame hidden]

\textbf{Prompt:}
{Now the system has completed a full turn. The self-reflection module believes that {\{model 1\}} is not capable of doing this job. Please return to the Reasoning and Retrieval stage and re-screen other models. The system has automatically updated your previous query results. Please go through Stage I.}

\end{tcolorbox}

This mechanism ensures that when initial model selections prove insufficient, the framework can systematically explore alternative candidates while maintaining the context of previously evaluated models, thereby improving the overall success rate of model discovery.

\section{Data Construction and Details}
\label{sec:Data}

\subsection{Model Cards}

This section describes the construction process of model cards for the 1,110 HuggingFace models in our dataset. Each model card is systematically organized to capture essential metadata and functional descriptions that enable effective model selection and retrieval.

% \subsubsection{Model Card Structure}
\paragraph{Model Card Structure.}

Following a similar approach to HuggingGPT, we construct model cards by extracting and organizing information from the HuggingFace model hub. Each card contains structured metadata including model type, supported tasks, performance metrics, and detailed descriptions. The standardized format ensures consistent information representation across all models in our database.

This is the JSON format of the specific model card data:

% \textbf{Example of Model Card Data in JSON Format:}

% \begin{verbatim}
% {
%   "id": "Model's ID", 
%   "task": "Model's task label", 
%   "downloads": "Model downloads", 
%   "likes": "Number of model likes", 
%   "meta": {
%     "language": ["Supported languages"], 
%     "datasets": ["Datasets used"], 
%     "license": "Model's license", 
%     "tags": ["List of model tags"],
%         ...
%   }, 
%   "description": "Model description"
%   ...
% }
% \end{verbatim}

\begin{tcolorbox}
[enhanced, colback=gray!4, colframe=gray!100, sharp corners, 
leftrule={0pt}, rightrule={0pt}, toprule={0pt}, bottomrule={0pt}, 
left={7pt}, right={2pt}, top={3pt}, bottom={3pt},
overlay={\draw[gray!100, line width=2.1pt] ([xshift=1.0pt]frame.north west) -- ([xshift=1.0pt]frame.south west);
         \draw[gray!70, line width=2pt] ([xshift=3pt]frame.north west) -- ([xshift=3pt]frame.south west);}, frame hidden]

\textbf{JSON:}
\begin{verbatim}
{
  "id": "Model's ID", 
  "task": "Model's task label", 
  "downloads": "Model downloads", 
  "likes": "Number of model likes", 
  "meta": {
    "language": Supported languages, 
    "datasets": Datasets used", 
    "license": "Model's license", 
    "tags": ["List of model tags"],
    ...
  }, 
  "description": "Model description"
  ...
}
\end{verbatim}

\end{tcolorbox}

% \subsubsection{Task Categories Coverage}
\paragraph{Task Categories Coverage.}

Our model cards encompass 37 distinct task categories, covering a comprehensive range of machine learning applications from natural language processing to computer vision and audio processing. Table~\ref{data} provides a complete overview of all supported tasks in our dataset, demonstrating the diversity and breadth of models available for selection.

% \subsection{Single-Task User Requests}
\paragraph{Single-Task User Requests.}

We systematically constructed 1,016 single-task user requests based on the characteristics of selected model cards from the HuggingFace repository. The construction process involved analyzing the capabilities and specializations of various models across different domains and task types. For each task category, we generated approximately 30 user requests, ensuring comprehensive coverage of different use cases, complexity levels, and domain-specific requirements. This approach allows us to evaluate our framework's performance across diverse scenarios that users might encounter in practice.

The user requests were crafted to vary in specificity and complexity, ranging from straightforward task descriptions (e.g., "I need a model for sentiment analysis") to more nuanced requirements that include domain constraints, performance preferences, or specific technical requirements (e.g., "I need a lightweight sentiment analysis model optimized for financial text that can run on mobile devices").

% \subsubsection{Annotation Schema}
% \paragraph{Annotation Schema.}

% Our dataset employs a dual-label annotation system performed by human experts to ensure high-quality ground truth labels:

% \noindent\textbf{1) Workability:} measures the likelihood that the selected model meets the basic user needs and functions properly. This metric reflects the system's ability to perform coarse-grained model filtering, essentially answering whether the recommended model can accomplish the requested task at a fundamental level. A model receives a positive workability score if it possesses the core functionality required by the user's request.

% \noindent\textbf{2) Reasonability:} is a stricter metric that requires the selected model not only to work correctly but also to align with the user's specific domain requirements or special constraints. This metric best captures the system's ability to select models that match the user's fine-grained preferences, considering factors such as model size, inference speed, domain specialization, language support, or other nuanced requirements specified in the user request.

% The dual-label system allows for nuanced evaluation: a model might be workable for a task but not reasonable given specific constraints, or vice versa. This granular assessment provides deeper insights into the framework's performance across different levels of selection precision.

% \subsubsection{Data Structure}
\paragraph{Data Structure.}
Each single-task user request in our dataset follows a structured format:

\begin{tcolorbox}
[enhanced, colback=gray!4, colframe=gray!100, sharp corners, 
leftrule={0pt}, rightrule={0pt}, toprule={0pt}, bottomrule={0pt}, 
left={7pt}, right={2pt}, top={3pt}, bottom={3pt},
overlay={\draw[gray!100, line width=2.1pt] ([xshift=1.0pt]frame.north west) -- ([xshift=1.0pt]frame.south west);
         \draw[gray!70, line width=2pt] ([xshift=3pt]frame.north west) -- ([xshift=3pt]frame.south west);}, frame hidden]

\textbf{JSON:}
\begin{verbatim}
{
  "request": "User Request", 
  "Task_label": {
    "workable": [Workable models], 
    "reasonable": [Reasonable models]
  }
}
\end{verbatim}

\end{tcolorbox}

\subsection{Multi-Task User Requests}

To better reflect real-world scenarios where users often have complex, multi-faceted requirements, we constructed multi-task user requests by leveraging GPT-4o to combine single-task requests into comprehensive scenarios. The ground truth labels for these multi-task requests are directly derived from the corresponding Single-Task User Requests labels:

\begin{tcolorbox}
[enhanced, colback=gray!4, colframe=gray!100, sharp corners, 
leftrule={0pt}, rightrule={0pt}, toprule={0pt}, bottomrule={0pt}, 
left={7pt}, right={2pt}, top={3pt}, bottom={3pt},
overlay={\draw[gray!100, line width=2.1pt] ([xshift=1.0pt]frame.north west) -- ([xshift=1.0pt]frame.south west);
         \draw[gray!70, line width=2pt] ([xshift=3pt]frame.north west) -- ([xshift=3pt]frame.south west);}, frame hidden]

\textbf{Prompt:}
{I would like to generate a multi-task user request. Below are several single-task user requests. Combine them into a natural, multi-task user request.}  

{1. The order of the tasks is not fixed; you can arrange the tasks in a way that aligns with a human query.} 
{2. The combined request must be coherent and meaningful, with each task relying on the results of the previous task(s).}
{3. If it is not possible to form a natural request, output 'NO'.}  
    
{Below are the user requests to combine:}

{\{user requests 1\}-\{task 1\}, ...}

\end{tcolorbox}

The generated multi-task requests underwent rigorous review and refinement by domain experts to ensure quality and coherence. Through this process, we curated 13,383 multi-task user requests. 

\section{Annotation Protocol}
\label{app:annotation}

This appendix provides detailed annotation guidelines, inter-annotator 
agreement analysis, and example annotations for our evaluation metrics.

\subsection{Workability Annotation Guidelines}

A model is considered \textit{workable} if it satisfies all of the following criteria:

\begin{enumerate}
\setlength{\itemsep}{-0.3em}
    \item \textbf{Task Type Match:} The model's task type (extracted 
    from the \texttt{pipeline\_tag} field in model metadata) matches 
    the user request.     
    \item \textbf{Input/Output Compatibility:} The model accepts 
    the required input format and produces the expected output format.     
    \item \textbf{Executability:} The model is available and 
    executable on Hugging Face Hub or can be deployed locally. 
    This is verified through:
    \begin{itemize}
        \item API availability check (for Hugging Face models)
        \item Checkpoint availability (for local deployment)
        \item No reported execution errors in model card
    \end{itemize}
\end{enumerate}

\textbf{Annotation Decision Tree:}
\begin{itemize}
\setlength{\itemsep}{-0.3em}
    \item If any criterion is not satisfied $\rightarrow$ Workability = 0
    \item If all criteria are satisfied $\rightarrow$ Workability = 1
\end{itemize}

\subsection{Reasonability Annotation Guidelines}

Among workable models, a model is considered \textit{reasonable} 
if it satisfies both of the following criteria:

\begin{enumerate}
\setlength{\itemsep}{-0.3em}
    \item \textbf{Domain-Relevant Training Data:} The model is 
    trained on datasets relevant to the user's domain. This is 
    verified through the \texttt{datasets} field in model cards.   
    \item \textbf{Competitive Performance:} The model achieves 
    competitive performance within its domain, defined as ranking 
    within the top-5 reported metrics among candidate models. 
    % For example:
    % \begin{itemize}
    %     \item For object-detection: mAP on domain-specific benchmarks
    %     \item For text-classification: Accuracy or F1 score
    % \end{itemize}
\end{enumerate}

\textbf{Annotation Decision Tree:}
\begin{itemize}
\setlength{\itemsep}{-0.3em}
    \item If the model is not workable $\rightarrow$ Reasonability = 0
    \item If the model is workable but fails either criterion 
    $\rightarrow$ Reasonability = 0
    \item If the model is workable and satisfies both criteria 
    $\rightarrow$ Reasonability = 1
\end{itemize}

\section{Additional Results}

\subsection{Experimental setup}
\label{sec:setup}
This section provides additional implementation details that complement the experimental configuration described in the main paper.

% \subsubsection{LLM Configuration} 
\paragraph{LLM Configuration.}
For the reasoning LLM component in our HuggingR${^4}$ framework, we maintain default parameter settings including temperature, Top-K, and Top-P values to preserve the model's natural reasoning capabilities. In contrast, for all other LLM components involved in the pipeline, we set the temperature to 0 to ensure deterministic and consistent outputs across experimental runs.

% \subsubsection{Vector Database and Retrieval Framework}
\paragraph{Vector Database and Retrieval Framework.}
We implement our vector database using Chroma, which provides efficient storage and retrieval of model embeddings. The retrieval operations are orchestrated through the LangChain framework, enabling seamless integration between the reasoning components and the vector database queries.

These additional parameters, combined with the configuration details provided in Section 4.2 of the main paper (Multi-Query number $K=4$, text-embedding-3-large embedding model, sliding window parameters $N=3$ and $k=5$, failure tracing threshold $\theta$=80\%), constitute the complete experimental setup for our evaluation.

% Table \ref{tab:software_versions} summarizes the specific versions of key software components used in our experiments.

% \begin{table}[h]
% \centering
% \caption{Software versions used in experiments}
% \vspace{-6pt}
% \begin{tabular}{lc}
% \toprule
% \textbf{Component} & \textbf{Version} \\
% \midrule
% langchain & 0.3.26 \\
% langchain-chroma & 0.2.4 \\
% langchain-core & 0.3.69 \\
% langchain-openai & 0.3.28 \\
% openai & 1.97.0 \\
% \bottomrule
% \end{tabular}
% \vspace{-10pt}
% \label{tab:software_versions}
% \end{table}

\subsection{Tool Selection vs. Model Selection}
\label{app:vs}

While both traditional API tool selection and community model selection involve choosing appropriate solutions for user tasks, they operate under fundamentally different paradigms. Traditional tool selection handles discrete, non-overlapping function categories (weather API and email API) with complete structured metadata, where selection reduces to categorical matching. In contrast, community model selection faces four critical challenges:

\begin{itemize}
    \item \textbf{Massive capability overlap:} Thousands of models target identical tasks (e.g., 6,117 sentiment analysis models on HuggingFace), making differentiation challenging.
    
    \item \textbf{Subtle differentiation:} Models vary in domain adaptation, 
          architecture, training data, and language support, requiring fine-grained comparative reasoning.
    
    \item \textbf{Continuous quality spectrum:} Models offer varying degrees of suitability rather than binary applicability.
\end{itemize}

These differences manifest across three critical dimensions. \textbf{First, search space characteristics.} Tool selection operates on 10 to 100 curated tools with minimal overlap, enabling exhaustive enumeration. Model selection navigates over 100,000 models with massive redundancy, requiring multi-stage progressive filtering. \textbf{Second, context dependency.} Tool performance is largely context-independent: a weather API returns accurate data regardless of application domain. In contrast, model performance is highly context-dependent. For instance, an object detector trained on everyday scenes (e.g., COCO) may achieve strong performance on common objects but struggle with domain-specific targets like satellite imagery or industrial defects, where specialized models demonstrate superior accuracy. \textbf{Third, selection criteria.} Tool selection uses binary applicability checks: Does the tool belong to the target category? Are required parameters available? This reduces to boolean logic. Model selection requires multi-dimensional ranking that balances task alignment, domain relevance, language compatibility, architectural suitability, and deployment constraints under incomplete information, necessitating context-aware scoring with uncertainty handling.

\subsection{Failure Tracing and Backtracking.}

To evaluate the effectiveness of the failure tracing mechanism, we conduct ablation studies by varying the similarity threshold $\theta$ that triggers the traceback process. Table~\ref{tab:ablation_theta} presents the results across different threshold settings. We observe a consistent improvement trend as $\theta$ increases from 0\% to 80\%, with performance peaking at $\theta=80\%$ (92.03\% workability and 82.46\% reasonability). Disabling the mechanism ($\theta=0\%$) yields the lowest performance (88.09\%/77.66\%), while an overly strict threshold ($\theta=100\%$) leads to slight degradation (91.24\%/81.99\%), indicating that moderate thresholds best balance failure detection and false positive avoidance.

\begin{table}[t]
\centering
\small
\caption{Ablation study on the failure tracing threshold $\theta$ using HuggingR$^4$ with GPT-4o-mini and text-embedding-3-large.}
\label{tab:ablation_theta}
\begin{tabular}{c|cc}
\toprule
\textbf{Threshold} $\theta$ & \textbf{Workability (\%)} & \textbf{Reasonability (\%)} \\
\midrule
0\%      & 88.09 & 77.66 \\
40\%               & 90.45 & 80.41 \\
60\%               & 91.73 & 81.69 \\
\textbf{80\%} & \textbf{92.03} & \textbf{82.46} \\
100\%              & 91.24 & 81.99 \\
\bottomrule
\end{tabular}
\end{table}

\subsection{Multi-Query Retrieval Strategy}

As mentioned in the main text, we employ multiple queries from different perspectives using LLMs before performing retrieval to achieve better results. However, unlike previous methods, we do not perform separate retrievals for each generated query. Instead, we concatenate these queries and conduct a single retrieval operation to achieve cost savings.

As shown in Table~\ref{tab:multi-query}, we conducted additional experiments to validate our approach. We compare our concatenation-based retrieval strategy against separate retrieval methods with different top-$k$ configurations using Reciprocal Rank Fusion (RRF) for result merging.

The results demonstrate that our concatenation approach achieves the best performance across both metrics. For the separate retrieval methods, we observe an interesting trade-off: as the top-$k$ value increases, workability improves while reasonability decreases. This phenomenon occurs because larger $k$ values expand the retrieval scope, capturing more potentially relevant models and improving workability. However, this broader scope introduces more noise and less precise candidates, which negatively impacts the subsequent fine-grained refinement process, leading to decreased reasonability scores.

\begin{table}[t]
\centering
\small
\caption{Comparison of Retrieval Strategies on HuggingR${^{4*}}$ with GPT-4o-mini and text-embedding-3-large.}
\begin{tabular}{lcc}
\toprule
\textbf{Method} & \textbf{Workability} & \textbf{Reasonability} \\
\midrule
Concatenation Retrieval & \textbf{84.77} & \textbf{75.56} \\
Separate Retrieval (Top-1) & 83.07 & 74.11 \\
Separate Retrieval (Top-3) & 83.66 & 73.72 \\
Separate Retrieval (Top-5) & 84.65 & 73.62 \\
\bottomrule
\end{tabular}
\vspace{-10pt}
\label{tab:multi-query}
\end{table}

\begin{table*}[th]
\caption{Overview of the task list, including arguments, candidate model pool sizes, and the number of single-task requests for each task.}
\label{tab:task_distribution}
\begin{center}
\resizebox{1.0\textwidth}{!}{
\begin{tabular}{l c c c}
        \toprule
        {\textbf{Task}} & {\textbf{Args}} & \textbf{\#Candidate Models} & \textbf{\#Single-Task Requests}\\
        \midrule
         
        % text-classification & text & 30  \\
        % token-classification & text & 30\\
        % HuggingR$^{4}$ & \textbf{92.03} & 82.46\\
        % \midrule
text-classification & text & 30 & 31\\
token-classification & text & 30 & 30\\ 
text2text-generation & text & 30 & 30\\
summarization & text & 30 & 32\\
translation & text  & 30 & 18\\
conversational & text & 30 & 14\\
text-generation & text  & 30 & 11\\
sentence-similarity & text  & 30 & 31\\
tabular-classification & text  & 30 & 30\\
object-detection & image  & 30 & 42\\
image-classification & image  & 30 & 30\\
image-to-text & image+text  & 30 & 31\\
text-to-image & text+image  & 30 & 19\\
image-to-image & image  & 30 & 38\\
visual-question-answering & image+text & 30 & 22\\
document-question-answering & text & 30 & 30\\
image-segmentation & image & 30 & 32\\
depth-estimation & image & 30 & 12\\
text-to-video & text+image & 30 & 28\\
text-to-speech & text+audio & 30 & 30\\
automatic-speech-recognition & audio & 30 & 34\\
audio-to-audio & audio & 30 & 33\\
audio-classification & audio & 30 & 36\\
tabular-regression & text & 30 & 25\\
time-series-forecasting & text & 30 & 30\\
voice-activity-detection & audio & 30 & 18\\
text-to-audio & text+audio & 30 & 20\\
table-question-answering & text & 30 & 19\\
question-answering & text & 30 & 30\\
keypoint-detection & image & 30 & 24\\
image-to-3d & image & 30 & 20\\
text-to-3d &  text-image  & 30 & 22\\
video-classification & video & 30 & 30\\
visual-document-retrieval & image+text & 30 & 30 \\
audio-text-to-text & audio+text & 30  & 33\\
image-text-to-text & image+text & 30 & 40\\
video-text-to-text & video+text & 30 & 31\\    
        
\bottomrule
       
\end{tabular}
}
\end{center}
\label{data}
\end{table*}

\begin{table*}[!t]
    \centering
    \fontsize{9pt}{10pt}\selectfont
    \caption{An illustrative example demonstrating how our proposed HuggingR$^4$ framework processes a user request.}
    \begin{tabular}{p{1.00\linewidth}}
    \midrule
        \rowcolor{gray!20}\textbf{Example \#1} \\
    \midrule
        \textbf{Question:} \\
        Could you perform a sentiment analysis on the tweets provided in ./tweets.txt?\\
        % \textbf{Labeled Answer:} \\
        % no \\
    \midrule
        \rowcolor{gray!20}
        \textbf{Model Output by HuggingR${^4}$}\\
    \midrule

To perform sentiment analysis on tweets, we need to select a model that is specifically designed for sentiment analysis, ideally trained or suitable for analyzing social media data like tweets. Let's go through the selection process step by step.

\textbf{Initial Similarity Retrieval}

Since the user wants to perform sentiment analysis on tweets, I will start with a similarity search query focusing on sentiment analysis related to social media or tweets.  

% \texttt{<|begin\_similarity\_query|>  perform sentiment analysis on tweets<|end\_similarity\_query|>}
\noindent\makebox[\linewidth]{\rule{0.439\linewidth}{0.4pt}~{Direct Retrieval}~\rule{0.439\linewidth}{0.4pt}}

% -------------- Direct Retrieval --------------

% \begin{verbatim}
\texttt{<|begin\_similarity\_query|>}
perform sentiment analysis on tweets
\texttt{<|end\_similarity\_query|>}
% \end{verbatim}

% \begin{verbatim}
\texttt{<|begin\_similarity\_result|>}

[finiteautomata/bertweet-base-sentiment-analysis,
cardiffnlp/twitter-roberta-base-sentiment, 
cardiffnlp/twitter-roberta-base-sentiment-latest, 
j-hartmann/emotion-english-distilroberta-base,
siebert/sentiment-roberta-large-english]

\texttt{<|end\_similarity\_result|>}

\noindent\makebox[\linewidth]{\rule{1.0\linewidth}{0.4pt}}
% ------------------------------------------------

\textbf{Analyzing the Similarity Results}

The models returned from the Retrieval search are as follows:

1. finiteautomata/bertweet-base-sentiment-analysis

2. cardiffnlp/twitter-roberta-base-sentiment

3. cardiffnlp/twitter-roberta-base-sentiment-latest

4. j-hartmann/emotion-english-distilroberta-base

5. siebert/sentiment-roberta-large-english

These models are designed for sentiment analysis, and several of them are specifically trained on Twitter data, which is ideal for the user's requirement.

Next, let's check if these models are specifically trained on tweet datasets to filter for the best ones further. 

\textbf{Dataset Filtering}

Considering the user's need for sentiment analysis on tweets, I'll filter for models that mention a relevant dataset related to social media or tweets.

% models designed for sentiment analysis on Twitter or social media datasets

\noindent\makebox[\linewidth]{\rule{0.429\linewidth}{0.4pt}~{Metadata Retrieval}~\rule{0.429\linewidth}{0.4pt}}
% -------------- Metadata Retrieval --------------

\texttt{<|begin\_dataset\_query|>}

models designed for sentiment analysis on Twitter or social media datasets

\texttt{<|end\_dataset\_query|>}

% \begin{verbatim}
\texttt{<|begin\_dataset\_result|>}

This retrieval is invalid. Please refer to other search results.

\texttt{<|end\_dataset\_result|>}

\noindent\makebox[\linewidth]{\rule{1.0\linewidth}{0.4pt}}

\textbf{Candidate Analysis}

Since the dataset filtering did not yield trustworthy results, I will proceed with analyzing the candidates based on their known capabilities and descriptions from the initial similarity results. 

I have the following candidates from the similarity retrieval:

1. finiteautomata/bertweet-base-sentiment-analysis
2. cardiffnlp/twitter-roberta-base-sentiment`
3. cardiffnlp/twitter-roberta-base-sentiment-latest

I'll now get detailed descriptions of these models to see which one would best fit the user's need for analyzing the sentiment in tweets.

\noindent\makebox[\linewidth]{\rule{0.436\linewidth}{0.4pt}~{Refinement Tool}~\rule{0.436\linewidth}{0.4pt}}
% -------------- Refinement Tool --------------

\texttt{<|begin\_descriptions\_query|>}

[finiteautomata/bertweet-base-sentiment-analysis, cardiffnlp/twitter-roberta-base-sentiment, cardiffnlp/twitter-roberta-base-sentiment-latest]

\texttt{<|end\_descriptions\_query|>}

\texttt{<|begin\_descriptions\_result|>}

[Model card 1: ..., model card 2: ..., model card 3: ...]

\texttt{<|end\_descriptions\_result|>}

\noindent\makebox[\linewidth]{\rule{1.0\linewidth}{0.4pt}}

All three models are capable of performing sentiment analysis on tweets. However, the `cardiffnlp/twitter-roberta-base-sentiment-latest` model stands out due to the following reasons:

- It is trained on a significantly larger dataset of tweets (124 million).

- It has more downloads and likes, indicating greater usage and community trust.

- It provides a comprehensive classification of sentiments.

After a module of Reflection, based on the analysis of the descriptions and the specific needs for sentiment analysis on tweets, I recommend:

\texttt{\textbackslash boxed\{cardiffnlp/twitter-roberta-base-sentiment-latest\}}. \\

\bottomrule
% \midrule
\end{tabular}
    % \end{tabular}

    \label{exp1}
\end{table*}

\begin{table*}[!t]
    \centering
    \fontsize{9pt}{10pt}\selectfont
    \caption{An illustrative example demonstrating how our proposed HuggingR$^4$ framework processes a user request.}
    \begin{tabular}{p{1.00\linewidth}}
    \midrule
        \rowcolor{gray!20}\textbf{Example \#2} \\
    \midrule
        \textbf{Question:} \\
        I have an image of a research paper. Can you identify the layout elements such as images and text within it?\\
        % \textbf{Labeled Answer:} \\
        % no \\
    \midrule
        \rowcolor{gray!20}
        \textbf{Model Output by HuggingR${^4}$}\\
    \midrule

To address the user's request for identifying layout elements such as images and text within an image of a research paper, we need to identify a model suitable for document layout analysis or optical character recognition (OCR) focused on document structures.

\textbf{Identifying the model requirements}

- Type of request: Layout element identification in an image of a research paper.

- Constraints: The model should be capable of handling layout elements like images and text.

\textbf{Using the Direct Retrieval tool}

Let's start by retrieval for models that are relevant to document layout analysis.

\noindent\makebox[\linewidth]{\rule{0.439\linewidth}{0.4pt}~{Direct Retrieval}~\rule{0.439\linewidth}{0.4pt}}

% -------------- Direct Retrieval --------------

% \begin{verbatim}
\texttt{<|begin\_similarity\_query|>}\\
Identify layout elements such as images and text in an image of a research paper.
\texttt{<|end\_similarity\_query|>}
% \end{verbatim}

% \begin{verbatim}
\texttt{<|begin\_similarity\_result|>}

[omoured/YOLOv10-Document-Layout-Analysis, TahaDouaji/detr-doc-table-detection, keremberke/yolov8s-table-extraction, \\ Aryn/deformable-detr-DocLayNet, keremberke/yolov8m-table-extraction]

\texttt{<|end\_similarity\_result|>}

\noindent\makebox[\linewidth]{\rule{1.0\linewidth}{0.4pt}}
% ------------------------------------------------

\textbf{Dataset Filtering}

Next, we’ll look to filter these models based on their suitability for datasets related to document layout analysis. This usually includes datasets with annotated layouts that include text and images.

Since we’re targeting layout identification in research papers, I will describe the required dataset more specifically.

\noindent\makebox[\linewidth]{\rule{0.429\linewidth}{0.4pt}~{Metadata Retrieval}~\rule{0.429\linewidth}{0.4pt}}
% -------------- Metadata Retrieval --------------

\texttt{<|begin\_dataset\_query|>}

Models trained on document layout datasets with elements such as text and images, specifically those suitable for research papers.

\texttt{<|end\_dataset\_query|>}

% \begin{verbatim}
\texttt{<|begin\_dataset\_result|>}

[omoured/YOLOv10-Document-Layout-Analysis, TahaDouaji/detr-doc-table-detection, keremberke/yolov8s-table-extraction, \\ Aryn/deformable-detr-DocLayNet, keremberke/yolov8m-table-extraction]

\texttt{<|end\_dataset\_result|>}

\noindent\makebox[\linewidth]{\rule{1.0\linewidth}{0.4pt}}

\textbf{Detailed Descriptions of the Candidate Models}

Since we still have more than three candidates, I will get detailed descriptions of these models to narrow it down further.

\noindent\makebox[\linewidth]{\rule{0.436\linewidth}{0.4pt}~{Refinement Tool}~\rule{0.436\linewidth}{0.4pt}}
% -------------- Refinement Tool --------------

\texttt{<|begin\_descriptions\_query|>}

[omoured/YOLOv10-Document-Layout-Analysis, TahaDouaji/detr-doc-table-detection, keremberke/yolov8s-table-extraction]

\texttt{<|end\_descriptions\_query|>}

\texttt{<|begin\_descriptions\_result|>}

[Model card 1: ..., model card 2: ..., model card 3: ...]

\texttt{<|end\_descriptions\_result|>}

\noindent\makebox[\linewidth]{\rule{1.0\linewidth}{0.4pt}}

\textbf{Final Selection}

- omoured/YOLOv10-Document-Layout-Analysis is the most suitable model since it is specifically designed for document layout analysis, capable of detecting not only text but also images, making it well-suited to the user's needs.

- The other models focus primarily on table detection, which may not fully address the broader requirement of identifying various layout elements in a research paper.

\textbf{Self-Reflection}

The selected model, omoured/YOLOv10-Document-Layout-Analysis, meets all the user's requirements as it focuses on document layout analysis and is trained on a dataset relevant to research paper layouts.

After a module of self-reflection, here is my final answer:

\texttt{\textbackslash boxed\{omoured/YOLOv10-Document-Layout-Analysis\}}. \\

\bottomrule
% \midrule
\end{tabular}
    % \end{tabular}
    % \label{tab:2}
    
    \label{exp2}
\end{table*}

\begin{table*}[!t]
    \centering
    \fontsize{9pt}{9pt}\selectfont
    \caption{An illustrative example demonstrating how our proposed HuggingR$^4$ framework processes a user request.}
    \begin{tabular}{p{1.00\linewidth}}
    \midrule
        \rowcolor{gray!20}\textbf{Example \#3} \\
    \midrule
        \textbf{Question:} \\
        Could you help me generate a speech clone that conveys specific emotions from the text: ./emotion\_text.txt?\\
        % \textbf{Labeled Answer:} \\
        % no \\
    \midrule
        \rowcolor{gray!20}
        \textbf{Model Output by HuggingR${^4}$}\\
    \midrule

\textbf{Analyze the User's Request}

- The user wants to generate a speech clone capable of conveying specific emotions based on text found in ./emotion\_text.txt. This indicates a requirement for a model that can perform emotion recognition or emotion synthesis in speech.

To find relevant models, I'll start with a similarity search related to generating speech that conveys emotions.

\noindent\makebox[\linewidth]{\rule{0.439\linewidth}{0.4pt}~{Direct Retrieval}~\rule{0.439\linewidth}{0.4pt}}

% -------------- Direct Retrieval --------------

% \begin{verbatim}
\texttt{<|begin\_similarity\_query|>}Speech synthesis that conveys emotions.
\texttt{<|end\_similarity\_query|>}
% \end{verbatim}

% \begin{verbatim}
\texttt{<|begin\_similarity\_result|>}

[SWivid/F5-TTS, SparkAudio/Spark-TTS-0.5B, HKUSTAudio/Llasa-3B, coqui/XTTS-v2, metavoiceio/metavoice-1B-v0.1]

\texttt{<|end\_similarity\_result|>}

\noindent\makebox[\linewidth]{\rule{1.0\linewidth}{0.4pt}}
% ------------------------------------------------

\textbf{Analyze Results}

The similarity retrieval returned 5 models. Now, I'll check if any of these models provide capabilities specific to synthesizing speech that conveys emotions.

\textbf{Filtering by Dataset}

Next, some of these models might have used datasets related to emotional speech synthesis. I'll filter based on that.

\noindent\makebox[\linewidth]{\rule{0.429\linewidth}{0.4pt}~{Metadata Retrieval}~\rule{0.429\linewidth}{0.4pt}}
% -------------- Metadata Retrieval --------------

\texttt{<|begin\_dataset\_query|>}

Models that utilize datasets for emotional speech synthesis.

\texttt{<|end\_dataset\_query|>}

% \begin{verbatim}
\texttt{<|begin\_dataset\_result|>}

This retrieval is invalid. Please refer to other search results.

\texttt{<|end\_dataset\_result|>}

\noindent\makebox[\linewidth]{\rule{1.0\linewidth}{0.4pt}}

The dataset filtering returned no valid models, indicating that I cannot trust this retrieval for emotional speech synthesis datasets. Therefore, I'll proceed directly with the candidate models obtained from the similarity search without additional filtering.

\textbf{Retrieve Descriptions of Candidate Models}

I will now get the detailed descriptions of the returned models to determine their suitability for generating emotional speech. 

\noindent\makebox[\linewidth]{\rule{0.436\linewidth}{0.4pt}~{Refinement Tool}~\rule{0.436\linewidth}{0.4pt}}
% -------------- Refinement Tool --------------

\texttt{<|begin\_descriptions\_query|>}

[SWivid/F5-TTS, SparkAudio/Spark-TTS-0.5B, HKUSTAudio/Llasa-3B]

\texttt{<|end\_descriptions\_query|>}

\texttt{<|begin\_descriptions\_result|>}

[Model card 1: ..., model card 2: ..., model card 3: ...]

\texttt{<|end\_descriptions\_result|>}

\noindent\makebox[\linewidth]{\rule{1.0\linewidth}{0.4pt}}

Given this analysis, I cannot confidently recommend a model that fully meets all requirements for generating emotionally expressive speech.

\textbf{Reflection}

Proceeding with the self-reflection on the models, since none explicitly state the ability to express specific emotions.

Now system have completed a full turn, The Reflection module believes that [SWivid/F5-TTS, SparkAudio/Spark-TTS-0.5B, HKUSTAudio/Llasa-3B] are not capable of doing this job. Please return to the thinking-retrieval stage and re-screen other models. The system has automatically updated your previous query results for you. Please go through the Step 1.

\noindent\makebox[\linewidth]{\rule{0.439\linewidth}{0.4pt}~{Direct Retrieval}~\rule{0.439\linewidth}{0.4pt}}

% \begin{verbatim}
\texttt{<|begin\_similarity\_query|>}Speech synthesis that conveys emotions.
\texttt{<|end\_similarity\_query|>}
% \end{verbatim}

% \begin{verbatim}
\texttt{<|begin\_similarity\_result|>}

[coqui/XTTS-v2, metavoiceio/metavoice-1B-v0.1, amphion/MaskGCT, lj1995/GPT-SoVITS, OuteAI/OuteTTS-0.1-350M]

\texttt{<|end\_similarity\_result|>}

\noindent\makebox[\linewidth]{\rule{1.0\linewidth}{0.4pt}}

Now, I will get detailed descriptions of these selected models for a final evaluation. 

\noindent\makebox[\linewidth]{\rule{0.436\linewidth}{0.4pt}~{Refinement Tool}~\rule{0.436\linewidth}{0.4pt}}
% -------------- Refinement Tool --------------

\texttt{<|begin\_descriptions\_query|>}

[coqui/XTTS-v2, metavoiceio/metavoice-1B-v0.1, lj1995/GPT-SoVITS]

\texttt{<|end\_descriptions\_query|>}

\texttt{<|begin\_descriptions\_result|>}

[Model card 1: ..., model card 2: ..., model card 3: ...]

\texttt{<|end\_descriptions\_result|>}

\noindent\makebox[\linewidth]{\rule{1.0\linewidth}{0.4pt}}

Considering the explicit emphasis on emotional synthesis and ease of use in various speech scenarios, I will select metavoiceio/metavoice-1B-v0.1 as the best fit due to its dedicated focus on emotional speech and strong specification in processing long-form text, which aligns well with the user's goal of generating expressive speech from a text file.

After a module of self-reflection, here is my final answer:
\texttt{\textbackslash boxed\{metavoiceio/metavoice-1B-v0.1\}}. \\

\bottomrule
% \midrule
\end{tabular}
    % \end{tabular}
    % \label{tab:3}
    
    \label{exp3}
\end{table*}

\subsection{Qualitative Results}

To demonstrate the effectiveness of HuggingR$^4$ in real-world, we present three representative examples that showcase different aspects of our framework's capabilities.

\paragraph{Example \#1.}
As shown in Table~\ref{exp1}, HuggingR$^4$ demonstrates its ability to understand domain-specific requirements. The system effectively recognizes that the user needs a sentiment analysis model specifically optimized for social media content. Through the reasoning-retrieval phase, our framework identifies that tweet sentiment analysis requires models trained on informal language, hashtags, and social media conventions, which differ significantly from traditional text sentiment analysis. The retrieved candidate models are then refined based on their training data sources, with preference given to models that explicitly mention Twitter or social media datasets in their descriptions.

% \textbf{Example \#2:}
\paragraph{Example \#2.}
As shown in Table~\ref{exp2}, this case exemplifies HuggingR$^4$'s capability to recognize multi-modal requirements and domain specificity simultaneously. The system correctly identifies that the user requires a computer vision model capable of document layout analysis, specifically tailored for academic papers. Through progressive reasoning, our framework understands that academic documents have unique structural elements that require specialized recognition capabilities. The refinement phase successfully filters models based on their training on academic document datasets and their ability to distinguish between different layout components typical in research papers.

\paragraph{Example \#3.}
As shown in Table~\ref{exp3}, This example particularly highlights the power of our reflection mechanism. Initially, the first round of retrieval returned suboptimal results, focusing primarily on generic text-to-speech models without emotional expressiveness capabilities. However, our reflection module identified this mismatch between user requirements and retrieved candidates. Recognizing the need for models with emotional speech synthesis capabilities, the system performed a sliding window backtrack and initiated a second retrieval round with refined search criteria. In the second iteration, HuggingR$^4$ successfully identified models specifically designed for emotional speech generation, demonstrating how the reflection component enables self-correction through explicit error analysis and iterative improvement in model selection.

%%%%%%%%%%%%%%%%%%%%%%%%%%%%%%%%%%%%%%%%%%%%%%%%%%%%%%%%%%%%%%%%%%%%%%%%%%%%%%%

\end{document}